%% file: template.tex
\documentclass{vgtc}           

\graphicspath{{figures/}{pictures/}{images/}{./}} 

\usepackage{times}                     
\newcommand{\hot}[1]{{\color{black} #1}}
\newcommand{\pin}[1]{{\color{black} #1}}
\usepackage{tabu}                      
\usepackage{booktabs}                  
\usepackage{multirow}                  
\usepackage{lipsum}                    
\usepackage{mwe}                       

\usepackage{mathptmx}                  
\usepackage{amsfonts}
\usepackage{bm}                        

\onlineid{1223}

\vgtccategory{IEEE VIS Short Paper}

\title{Lossless-INR: Lossless Volumetric Implicit Neural Representations}

\author{Kaiyuan Tang\thanks{e-mail: ktang2@nd.edu} %
\and Daniel Burke\thanks{e-mail: dburke6@nd.edu} %
\and Chaoli Wang\thanks{e-mail: chaoli.wang@nd.edu}}
\affiliation{\scriptsize University of Notre Dame}

\authorfooter{
  \item  
  The authors are with
the Department of Computer Science and Engineering, University of Notre Dame, Notre Dame, IN 46556, USA.\\
E-mail: \{dburke6, ktang2, chaoli.wang\}@nd.edu.
}

\teaser{
  \centering
  \setlength{\tabcolsep}{2pt}
  \renewcommand{\arraystretch}{1.0}
  \begin{tabular}{@{}ccccc@{}}
    {\scriptsize model size / PSNR} &
    {\scriptsize 12.5\,MB / $\bm{+}${\normalsize $\bm{\infty}$}\,dB} &
    {\scriptsize 67.9\,MB / 52.5\,dB} &
    {\scriptsize 13.7\,MB / 47.6\,dB} &
    {\scriptsize 1.7\,MB / 42.3\,dB} \\
    \includegraphics[width=0.19\linewidth]{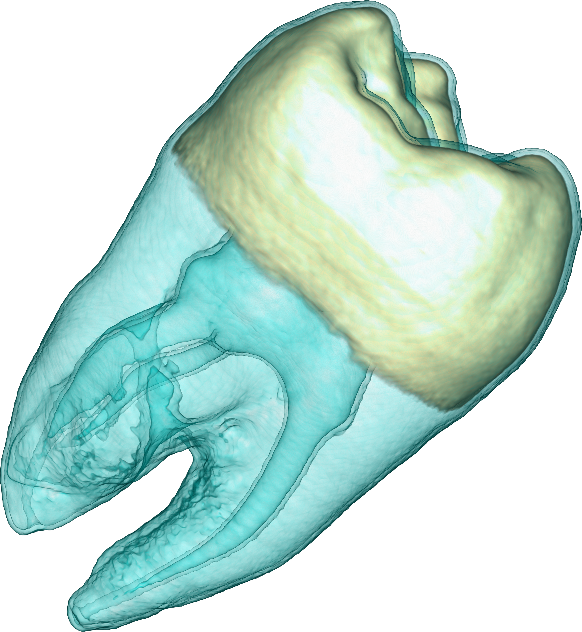} &
    \includegraphics[width=0.19\linewidth]{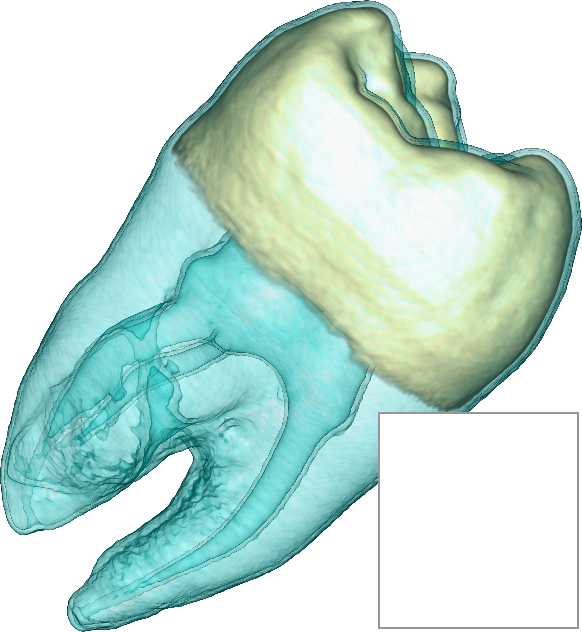} &
    \includegraphics[width=0.19\linewidth]{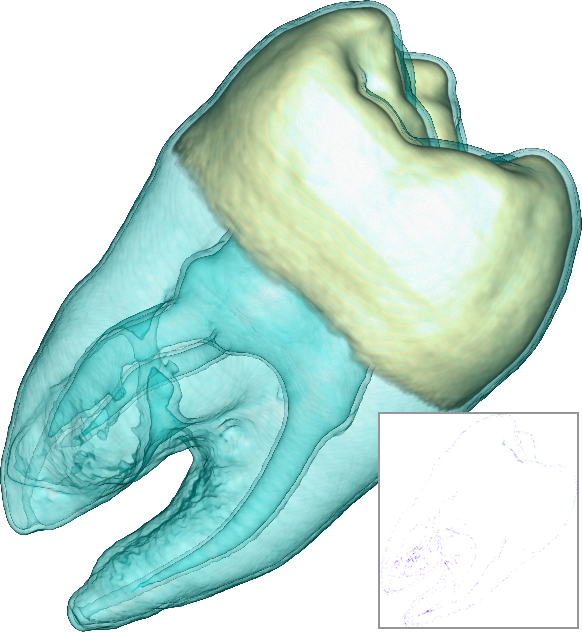} &
    \includegraphics[width=0.19\linewidth]{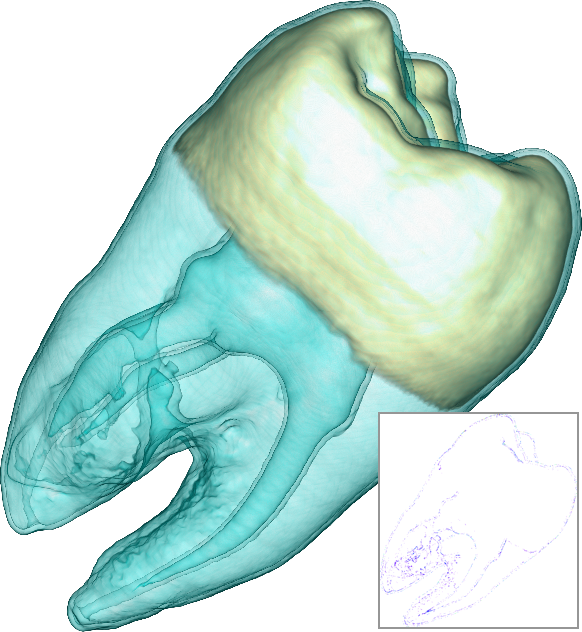} &
    \includegraphics[width=0.19\linewidth]{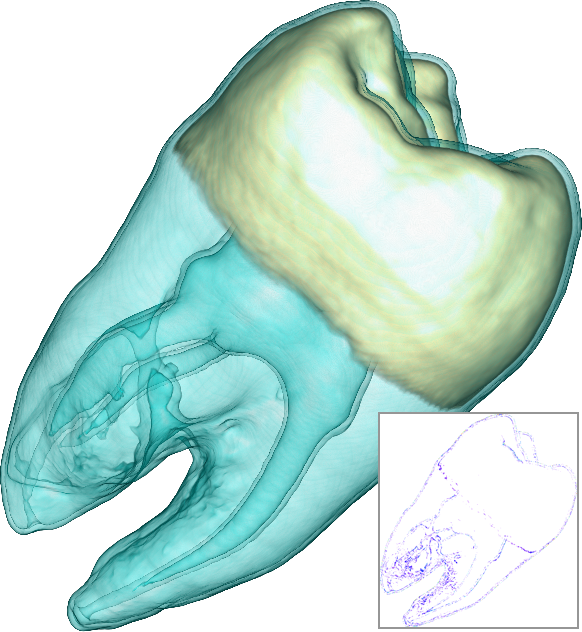} \\
    {\scriptsize ground truth} &
    {\scriptsize Lossless-INR} &
    {\scriptsize \pin{Instant-NGP (large)}} &
    {\scriptsize \pin{Instant-NGP (medium)}} &
    {\scriptsize \pin{Instant-NGP (small)}} \\
  \end{tabular}
  \vspace{-0.125in}
  \caption{%
    Comparison of the rendering results on the tooth dataset using Instant-NGP~\cite{Muller-TOG22} under different model sizes and our Lossless-INR. The bottom-right difference image shows pixel-wise error relative to the ground truth in the CIELUV color space.
In this paper, we explore and design Lossless-INR, which can represent volumetric data without error while maintaining a relatively compact model size.}
\label{fig:motivation}
}

\abstract{
Implicit neural representation (INR) methods provide continuous coordinate-to-value mappings and integrate naturally with direct volume rendering, making them attractive for representing volumetric data. However, existing INR-based approaches for volumetric data are inherently lossy, and even small reconstruction errors can propagate through rendering and downstream analysis. In this work, we explore Lossless-INR, a lossless INR framework for 3D scientific volumetric data based on bit-plane decomposition. By decomposing each voxel value into binary bit-planes, we reformulate reconstruction as per-bit binary classification, so that exact recovery reduces to predicting every bit correctly. To make this optimization tractable while keeping the representation compact, we combine an octree block-partitioning strategy that adaptively subdivides complex regions with a ternary feature-grid network whose grid entries are parameterized by a ternary set of values. Experiments on diverse volumetric datasets show that this design can achieve zero bit-error rate and bit-exact reconstruction, enabling faithful rendering and downstream analysis with a compact representation. The code is available at \url{https://github.com/TouKaienn/Lossless-INR}.
} 


\begin{document}


\vspace{-0.075in}
\firstsection{Introduction}

\maketitle

\input{intro}

\vspace{-0.075in}
\input{related}

\vspace{-0.075in}
\input{method}
\vspace{-0.075in}

\input{results}

\vspace{-0.075in}
\input{conclusions}

\vspace{-0.075in}
\acknowledgments{This research was supported in part by the U.S.\ National Science Foundation through grants IIS-2101696, OAC-2104158, IIS-2401144, and CCF-2550610. The authors thank the anonymous reviewers for their insightful comments.}

\vspace{-0.05in}
\bibliographystyle{abbrv-doi-hyperref-narrow}

\bibliography{template-abbv}

\appendix
\input{appendix}

\end{document}

%% file: intro.tex

Implicit neural representation (INR) models volumetric data as continuous coordinate-to-value functions parameterized by neural networks, supporting random-access queries and integrating naturally into direct volume rendering (DVR) pipelines.
Because of their simple architecture and flexibility, INRs have become a common choice for volume visualization, with applications spanning compression~\cite{Lu-CGF21, Tang-PacificVis24, Han-TVCG24, Han-TVCG26, Yang-PacificVis25}, rendering~\cite{Weiss-CGF22, Wu-TVCG24, Wurster-TVCG24, Wurster-PacificVis25}, data generation~\cite{Han-TVCG23, Tang-CG24}, scene representation~\cite{Yao-PVIS25, Yao-CG25, Tang-TVCG25}, and ensemble exploration~\cite{Chen-TVCG25}.

Despite their effectiveness, existing INR-based methods for volumetric data are inherently \emph{lossy}~\cite{Lu-CGF21, Weiss-CGF22, Tang-PacificVis24, Han-TVCG24}.
Even small reconstruction errors can propagate through multiple stages of the DVR pipeline, \pin{from} transfer function lookups to gradient estimation for shading, leading to incorrect colors and lighting during rendering.
These errors are also problematic for analysis tasks such as computing derived fields, extracting topology, or comparing simulation ensembles, where small reconstruction errors can significantly bias analysis results.
A straightforward solution is to increase model capacity by adding more optimizable parameters to reduce reconstruction error.
However, as shown in Figure~\ref{fig:motivation}, even a substantial increase in parameter budget does not eliminate the error, leaving a persistent gap from the original data.

To address this limitation, we design Lossless-INR, a lossless INR for 3D scientific volumetric data via \emph{bit-plane decomposition}~\cite{Punnappurath-TPAMI22, Han-CVPR23, Han-CVPR25}.
Instead of directly regressing continuous voxel values, we decompose each $B$-bit voxel value into $B$ binary bit-planes and reformulate reconstruction as per-bit binary classification, with network outputs quantized to $\{0,1\}$.
If every bit of every voxel is predicted correctly, the original voxel values can be recovered exactly by recomposing the predicted bit-planes, yielding a bit-exact lossless representation.
Applying this idea to 3D volumes, however, is non-trivial because driving the bit error to zero still requires substantial model capacity.
To keep the representation compact while making the learning problem tractable, we combine an \emph{octree block partitioning} strategy that adaptively divides complex volumes into smaller local blocks with a \emph{ternary feature grid network} whose grid entries are parameterized with a ternary set.
This design turns the full reconstruction problem into a collection of manageable local fitting tasks while preserving a compact and lossless overall representation.

The contributions of this paper are summarized as follows:
First, to our knowledge, we present the first INR framework that achieves lossless representation of volumetric data.
Second, we extend bit-plane decomposition to 3D volumes and combine it with an octree block partitioning strategy and a ternary feature grid network to obtain a storage-efficient lossless representation.
Third, through experiments on diverse volumetric datasets, we show that our Lossless-INR achieves zero bit-error rate at every voxel, enabling bit-exact reconstruction and rendering.

%% file: related.tex
\section{Related Work}

{\bf Deep learning for volume representation.}
As a core DL4SciVis research task~\cite{Wang-TVCG23}, INRs encode a volumetric scalar field as a continuous function learned by a neural network, enabling compressed storage and random-access evaluation.
Lu et al.~\cite{Lu-CGF21} framed volume compression as function approximation, showing that a coordinate-based multilayer perceptron (MLP) with quantized weights can outperform classical codecs while supporting time-varying fields and gradient preservation.
Weiss et al.~\cite{Weiss-CGF22} redesigned the network to exploit GPU tensor cores, integrating reconstruction directly into on-chip raytracing kernels for significantly faster decoding and compression-domain volume rendering.
Tang and Wang~\cite{Tang-PacificVis24} proposed ECNR, which replaces a single large MLP with a Laplacian pyramid of small, parallelized MLPs that fit local blocks, substantially accelerating training and inference for time-varying datasets.
Han et al.~\cite{Han-TVCG24} introduced KD-INR, applying knowledge distillation to transfer temporal redundancy across timesteps for more efficient encoding.
Tang and Wang~\cite{Tang-CG24} further presented STSR-INR, a method for simultaneous spatiotemporal super-resolution of multivariate volumes via learnable variable embeddings and latent-space interpolation.
Yang et al.~\cite{Yang-PacificVis25} proposed Meta-INR, a meta-learning pretraining strategy that learns generalizable initial parameters from partial observations, enabling rapid adaptation to unseen volumes.
Building on Meta-INR, Son et al.~\cite{Son-VISSP25} further developed MC-INR to encode multivariate scientific data efficiently. 
Wurster et al.~\cite{Wurster-TVCG24} developed APMGSRN, which adaptively places multi-resolution feature grids for large-scale data visualization.
Wu et al.~\cite{Wu-TVCG24} combined multi-resolution hash encoding with neural representations for interactive volume visualization.
These hybrid architectures, inspired by the multi-resolution hash encoding of M\"{u}ller et al.~\cite{Muller-TOG22}, have proven particularly effective in balancing reconstruction quality, model compactness, and rendering speed.
\hot{Beyond coordinate-based INRs, 3D Gaussian splatting has recently been explored as an alternative for explorable volume visualization~\cite{Tang-TVCG25b}, scene editing~\cite{Tang-TVCG26}, natural-language interaction~\cite{Ai-TVCG26}, feature tracking~\cite{Yao-VIS25}, compressive representation~\cite{Tang-VIS26}, and virtual reality application~\cite{Jeon-VIS26}.}

However, all of the above methods are fundamentally lossy and cannot guarantee exact reconstruction of the original data.
Unlike these approaches, we target lossless representation, ensuring that every voxel is reconstructed without any error.
Rather than minimizing a regression loss that tolerates residual error, our Lossless-INR reformulates fitting as per-bit binary classification via bit-plane decomposition for lossless volumetric data representation.

{\bf Lossless representation.}
Achieving lossless representation with neural networks requires the predicted output to match the original signal at every sample point with zero error.
For example, Punnappurath and Brown~\cite{Punnappurath-TPAMI22} showed that operating on individual bit-planes can recover lost bit-depth information, demonstrating the utility of bit-level signal decomposition.
Han et al.~\cite{Han-CVPR23} introduced ABCD, which adds a bit coordinate to the network input for arbitrary bitwise coefficient prediction.
Han et al.~\cite{Han-CVPR25} further presented a bit-plane decomposition method that reformulates signal fitting as per-bit binary classification, thereby reducing the theoretical upper bound on the number of parameters and enabling lossless INR for 2D images and audio, even at 16-bit precision.

However, directly applying lossless INR training to 3D volumetric data poses significant challenges: the substantially larger data size demands prohibitively large models, and convergence to zero error becomes increasingly difficult as the spatial dimensionality grows.
In this work, we address these challenges through ternary-set parameterization of feature-grid entries, which drastically reduces per-parameter storage cost, and a block-partitioning strategy that decomposes the volume into manageable subproblems, thereby making lossless convergence tractable for volumetric 3D data.

%% file: method.tex
\section{Method}

\begin{figure}[!h]
\centering
\includegraphics[width=\linewidth]{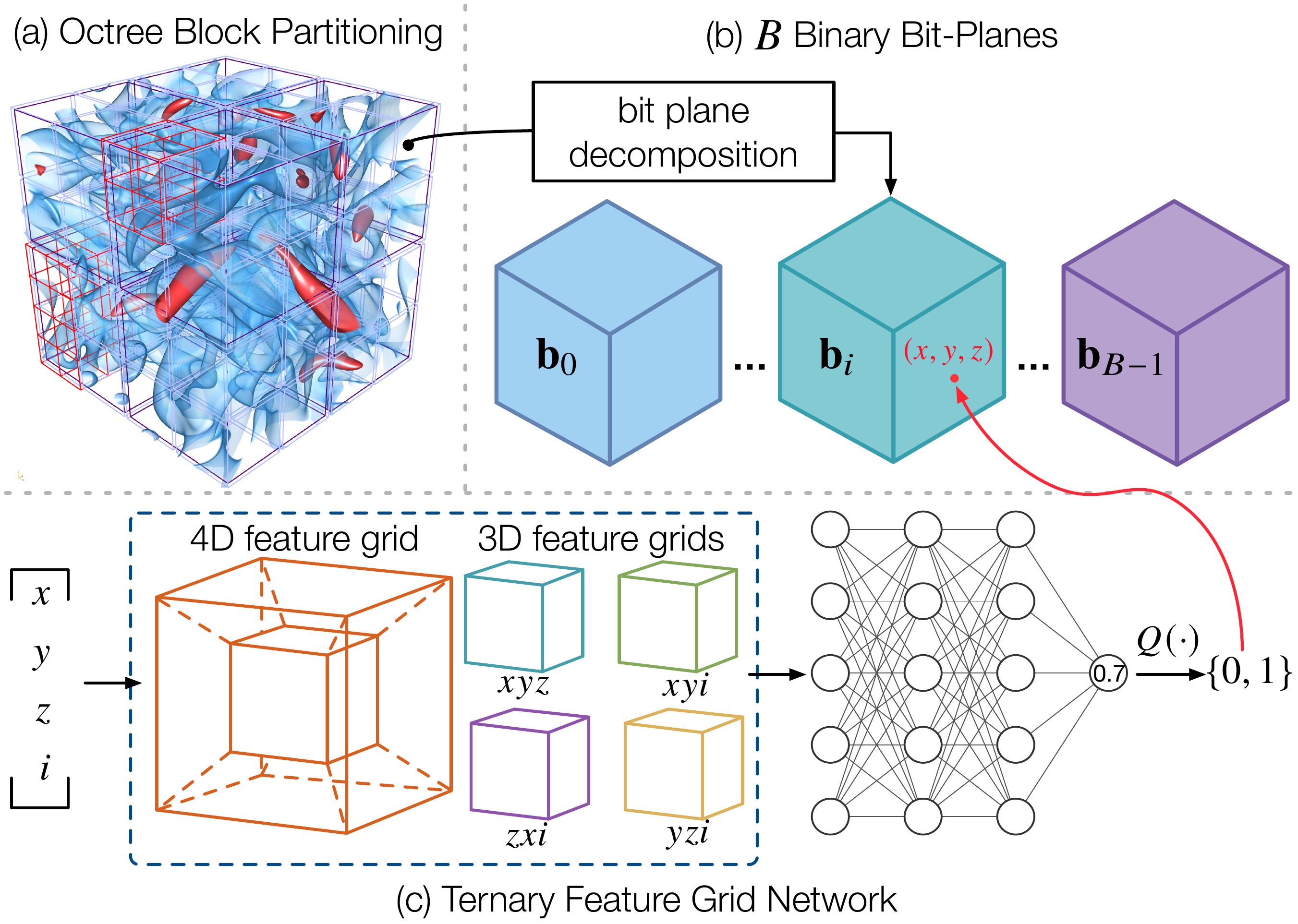}
\vspace{-.2in}
\caption{Overview of Lossless-INR. (a) The entire volume is adaptively divided into local blocks with an octree structure, and each block is decomposed into (b) $B$ binary bit-planes, (c) a ternary feature grid network is utilized to losslessly encode each block by classifying bit values.}
\label{fig:overview}
\vspace{-.1in}
\end{figure}

Lossless-INR achieves lossless volumetric representation through three components, as illustrated in Figure~\ref{fig:overview}: \emph{bit-plane decomposition}, \emph{octree block partitioning}, and a \emph{ternary feature grid network}.
Bit-plane decomposition (Section~\ref{subsec:bit-plane}) first decomposes volume data into a set of bit-planes, transforming voxel-value regression into binary classification and thereby enabling lossless representation. 
However, fitting all bit values over an entire volume remains difficult due to limited network capability. We therefore introduce octree block partitioning (Section~\ref{subsec:octree}) to adaptively divide complex volumes into simpler local blocks. Finally, we leverage a ternary feature grid network (Section~\ref{subsec:ternary}) whose feature values are constrained to $\{-1, 0, +1\}$ to encode each block, enabling lossless reconstruction while maintaining a compact model size.

\vspace{-0.05in}
\subsection{Bit-Plane Decomposition}
\label{subsec:bit-plane}

To overcome the inherent precision limitations of continuous voxel value regression, we reformulate the volumetric fitting problem as a binary classification task. Given a volumetric scalar field where each voxel value is represented by $B$ bits of precision, we decompose the voxel value into a set of $B$ binary bit-planes $\{\mathbf{b}_0, \mathbf{b}_1, \dots, \mathbf{b}_{B-1}\}$. 
For each bit-plane $\mathbf{b}_i$, we optimize the network $f_{\theta}$ as follows
\begin{equation}
\vspace{-0.025in}
\hat{\theta} = \arg\min_{\theta} \mathcal{L}\left(\mathbf{b}_i(x,y,z), f_{\theta}(x,y,z,i)\right),
\vspace{-0.025in}
\end{equation}
\hot{where $\left(x,y,z\right)$ indicates the spatial position, $\hat{\theta}$ denotes the optimized parameters, and $\mathcal{L}$ corresponds to the binary cross-entropy loss.
After training, let $\mathbf{V}(x,y,z)$ denote the voxel value at position $\left(x,y,z\right)$. The network reconstructs the volume data as}
\begin{equation}
\vspace{-0.025in}
\label{eqn:bit-equation}
\mathbf{V}(x,y,z) = \frac{1}{2^B - 1} \sum_{i=0}^{B-1} 2^i Q\Big(f_\theta(x,y,z,i)\Big),
\vspace{-0.025in}
\end{equation}
where $Q(\cdot)$ is the quantization function that maps the network's continuous output to the discrete bit value $\{0, 1\}$. 
If $f_{\theta}$ can predict every bit of target volume correctly, it can be considered as a lossless neural representation of that volume.

\vspace{-0.05in}
\subsection{Octree Block Partitioning}
\label{subsec:octree}

Fitting an entire scientific volume with a single network $f_{\theta}$ is not trivial; the complexity of optimization can grow quickly with the increase of both spatial resolution and bit precision. We therefore hierarchically divide the volume and encode each block with a separate network.
We start from a uniform partition of the volume into non-overlapping blocks and fit each block with its own $f_{\theta}$ for a fixed number of training iterations. If, at the end of training, $f_{\theta}$ predicts all $\mathbf{b}_i(x, y, z)$ correctly for every bit index $i \in \{0, \dots, B-1\}$, the block is accepted as an octree leaf. Otherwise, the block is spatially divided into smaller children, each of which becomes a new candidate block. The procedure repeats recursively until every leaf block can be fit losslessly.
\pin{In practice, we initialize the partition into blocks of size $64^3$ and recursively subdivide them down to a minimum block size of $16^3$.}
This adaptive scheme matches model capacity to the complexity of local content. However, because an individual network fits each leaf block, the total number of parameters grows rapidly with the number of leaves; this motivates the ternary feature grid network described next, which keeps each per-block network compact.

\vspace{-0.05in}
\subsection{Ternary Feature Grid Network}
\label{subsec:ternary}

Our per-block INR consists of multiple feature grids and a lightweight MLP decoder: the feature grids jointly map the input coordinates $(x, y, z, i)$, where $i$ indexes the bit plane, to a continuous spatial feature, and the MLP decodes the feature into a bit value. To keep each per-block model compact, every entry of the feature grids is constrained to a ternary set of parameters, while the MLP decoder stays at full precision.

{\bf Feature grids.}
We build the encoder on the multi-resolution hash-grid design of Instant-NGP~\cite{Muller-TOG22}, in which features are stored on hierarchical grids and queried through linear interpolation. Our encoder consists of one 4D hash grid that operates on the full coordinates $(x, y, z, i)$, together with four 3D hash grids, each operating on a distinct triplet of axes. For each query coordinate, we look up the feature vectors at each grid and concatenate them into a single spatial feature, which is then fed to the MLP for decoding. The 4D grid captures bit-aware spatial structure, while the 3D grids provide complementary lower-dimensional views to improve representation.

{\bf Ternary parameterization.}
Following recent extreme-low-bit network designs~\cite{Shin-NeurIPS23, Ma-arXiv24-2}, we constrain every entry of every hash grid to a ternary set during training. Given a grid weight tensor $\mathbf{W}$, we scale $\mathbf{W}$ by its mean absolute value $\gamma$, round each entry to the nearest integer in $\{-1, 0, +1\}$, and rescale back by $\gamma$, i.e.,
\begin{equation}
\label{eqn:ternary}
\vspace{-0.025in}
\widetilde{\mathbf{W}} \;=\; \gamma \cdot \mathrm{RoundClip}\!\left(\frac{\mathbf{W}}{\gamma + \epsilon},\, -1,\, 1\right), \quad \gamma \;=\; \frac{1}{n}\sum_{j} |W_j|,
\vspace{-0.025in}
\end{equation}
where $\mathrm{RoundClip}(x, a, b) = \max(a, \min(b, \mathrm{round}(x)))$, $n$ is the number of grid entries, and a small value $\epsilon$ for numerical stability. Each entry of resulting weight $\widetilde{\mathbf{W}}$ thus takes one of three values $\{-\gamma, 0, +\gamma\}$, so every hash-grid tensor is fully described by \pin{a ternary digit} in $\{-1, 0, +1\}$ per entry together with a single per-tensor scale $\gamma$. 
By optimizing and storing ternary weights rather than floating-point weights, our Lossless-INR can achieve a lossless representation while maintaining a relatively compact model size.

\begin{table}[!htb]
\centering
\caption{Volumetric datasets used in our evaluation.}
\label{tab:test_dataset}
\vspace{-0.05in}
\resizebox{0.8\columnwidth}{!}{%
\begin{tabular}{ccc}
\toprule
dataset & volume resolution ($x \times y \times z$) & data type \\
\midrule
\textsf{engine}         & 256$\times$256$\times$128             & uint8   \\
\textsf{foot}           & 256$\times$256$\times$256                & uint8   \\
\textsf{MRI-woman}      & 256$\times$256$\times$109             & uint16  \\
\textsf{tooth}          & 103$\times$94$\times$161              & uint8   \\
\textsf{vortex}          & 128$\times$128$\times$128             & float32 \\
\bottomrule
\end{tabular}
}
\vspace{-.1in}
\end{table}

\begin{figure*}[!t]
\centering
\setlength{\tabcolsep}{2pt}
\renewcommand{\arraystretch}{1.0}
\begin{tabular}{@{}cccccc@{}}
    \includegraphics[width=0.1575\linewidth]{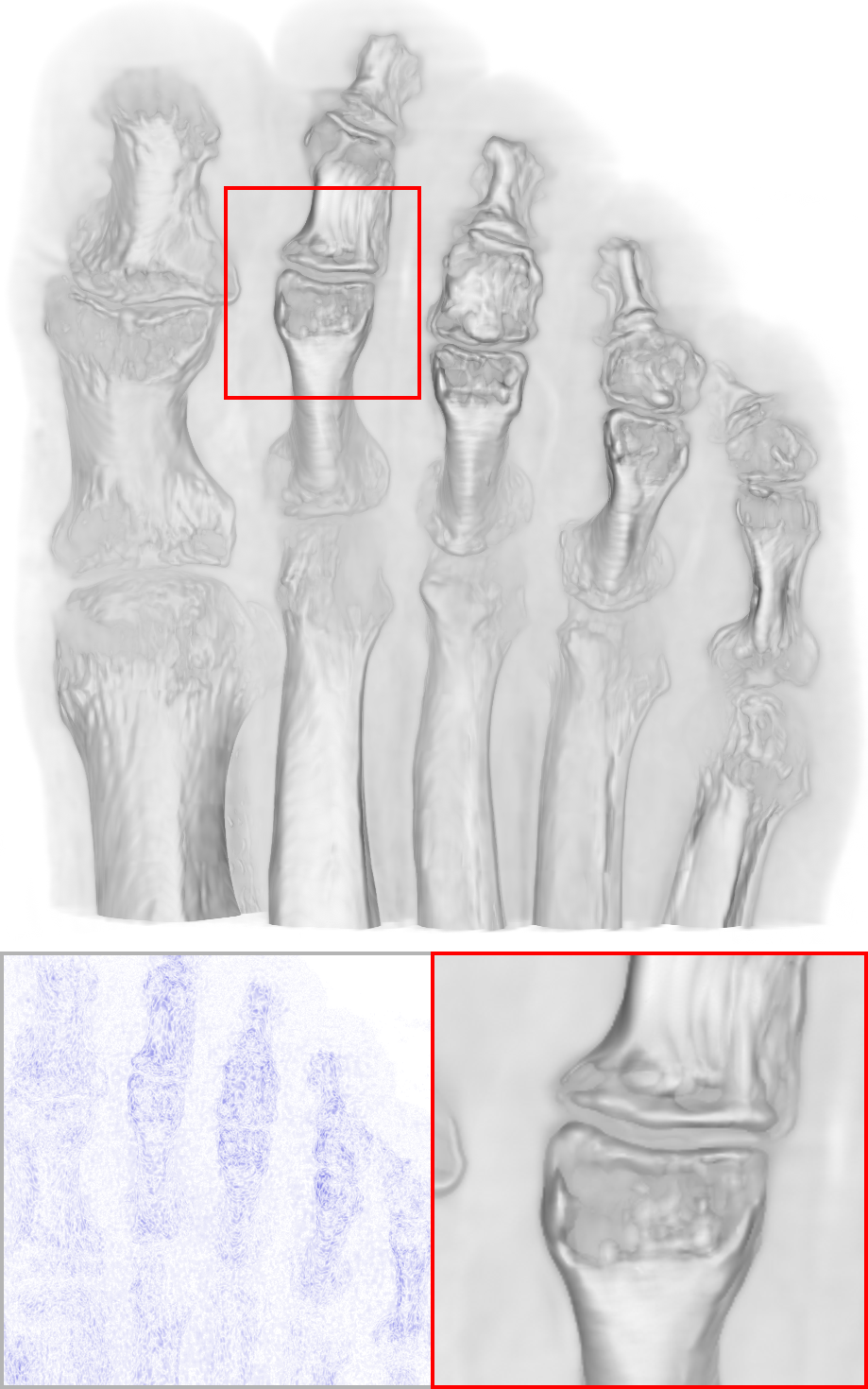} &
    \includegraphics[width=0.1575\linewidth]{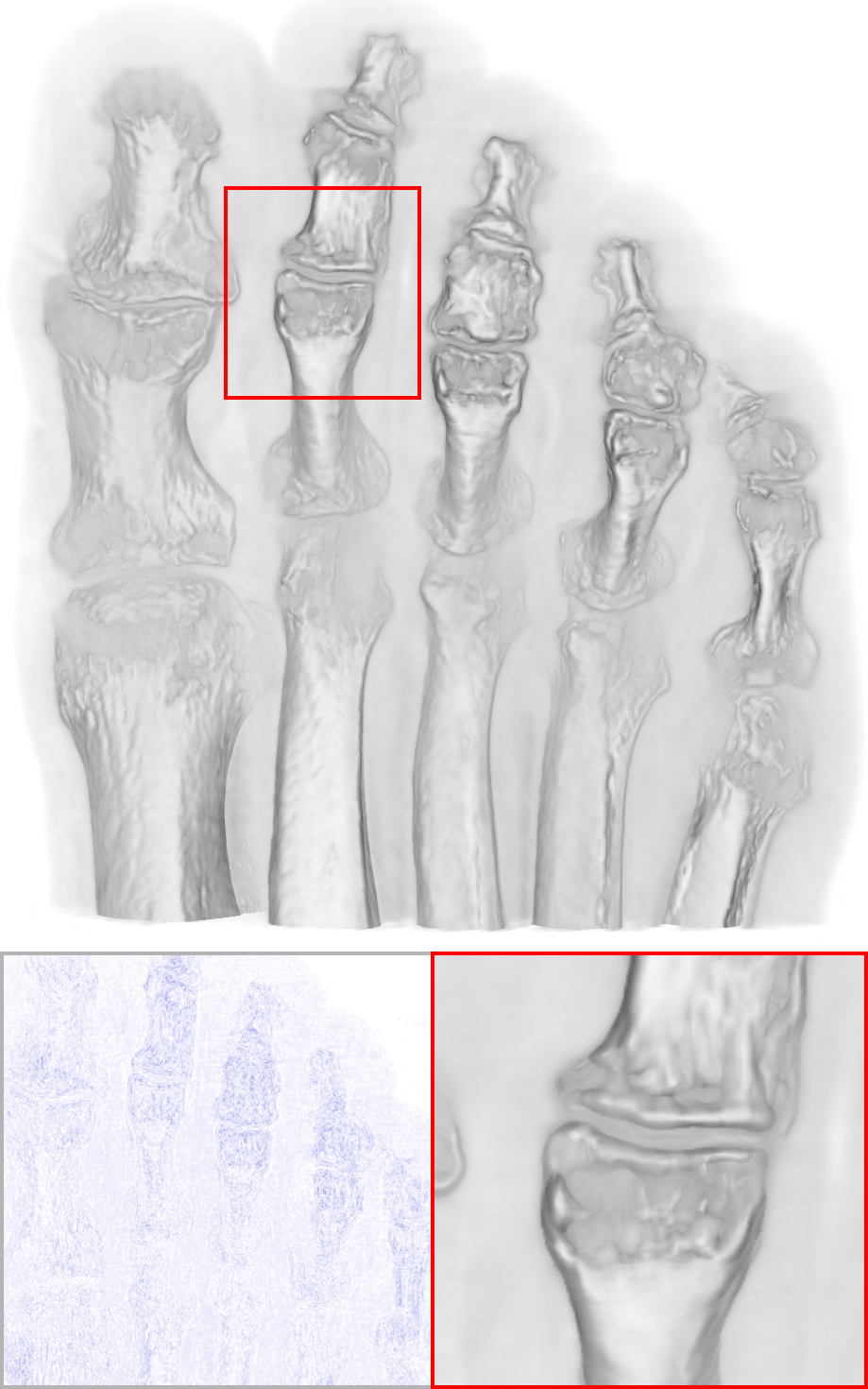} &
    \includegraphics[width=0.1575\linewidth]{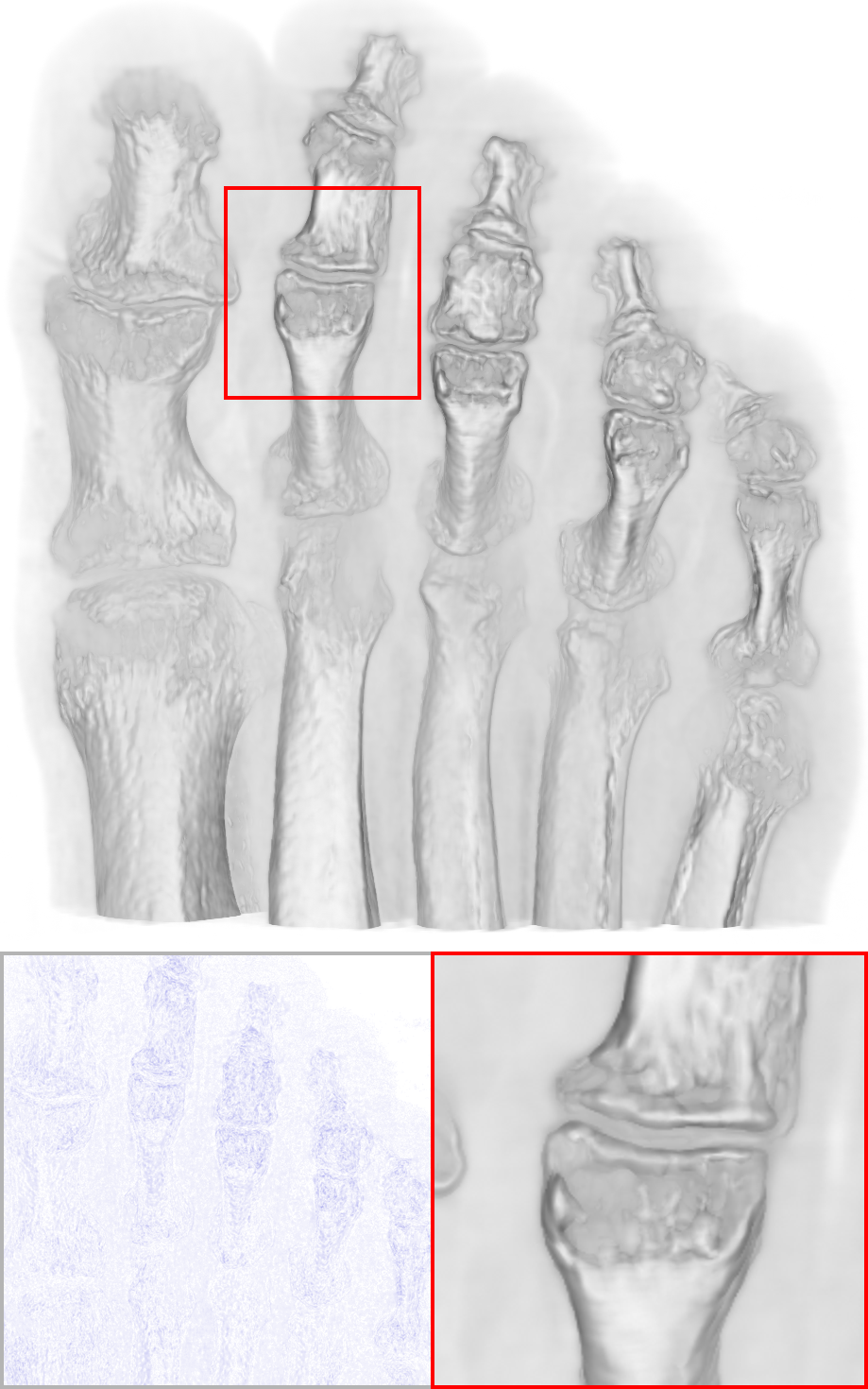} &
    \includegraphics[width=0.1575\linewidth]{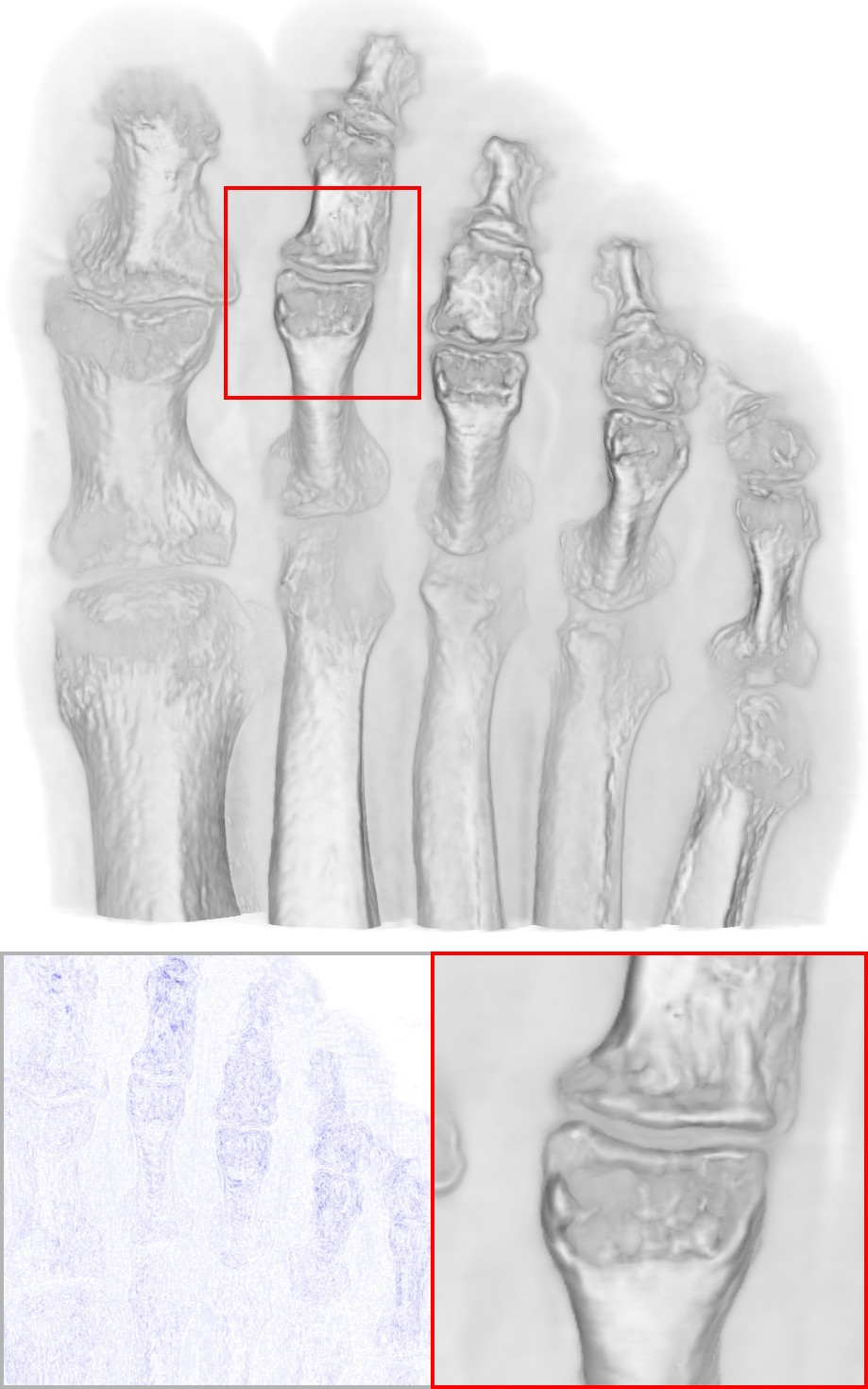} &
    \includegraphics[width=0.1575\linewidth]{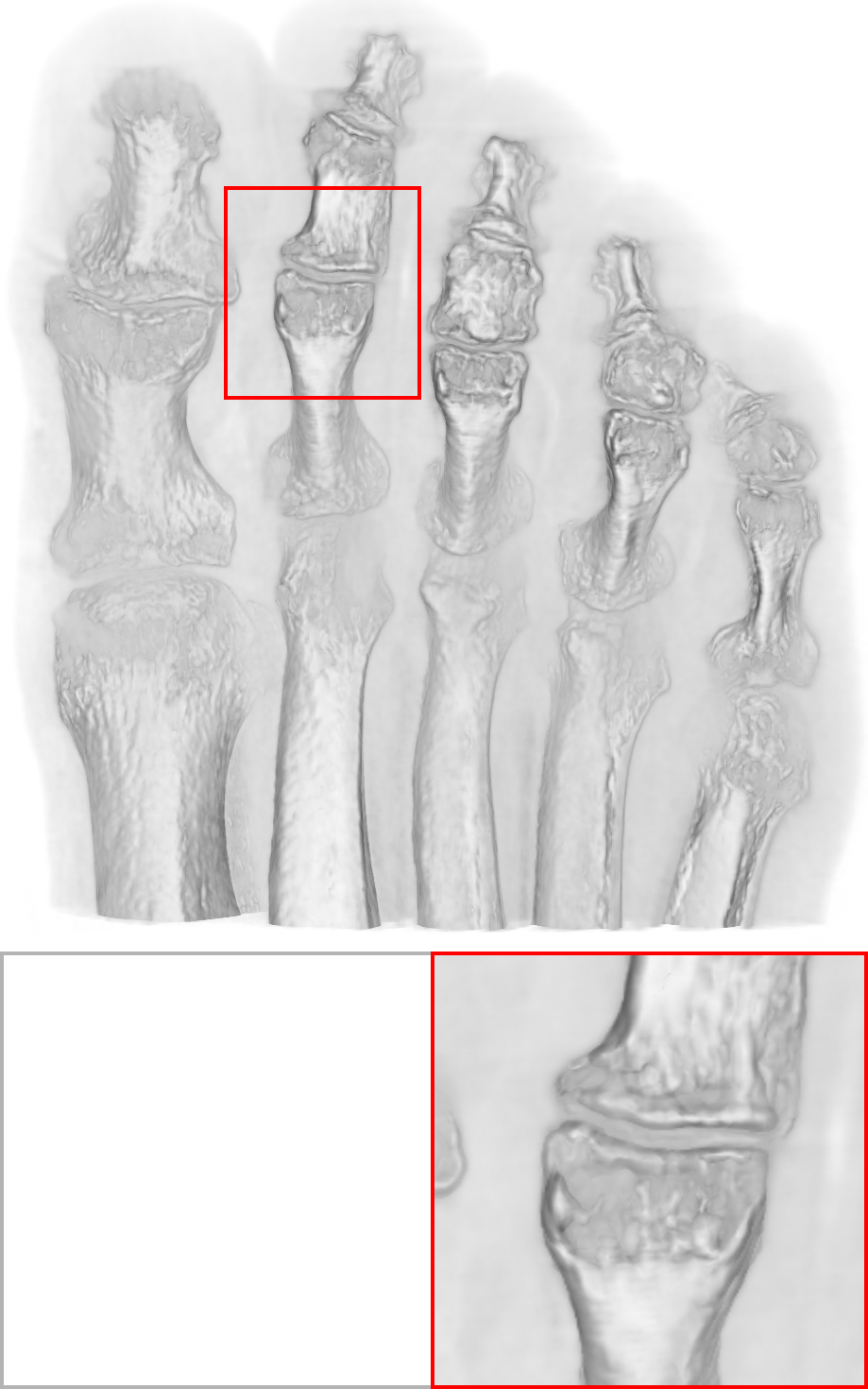} &
    \includegraphics[width=0.1575\linewidth]{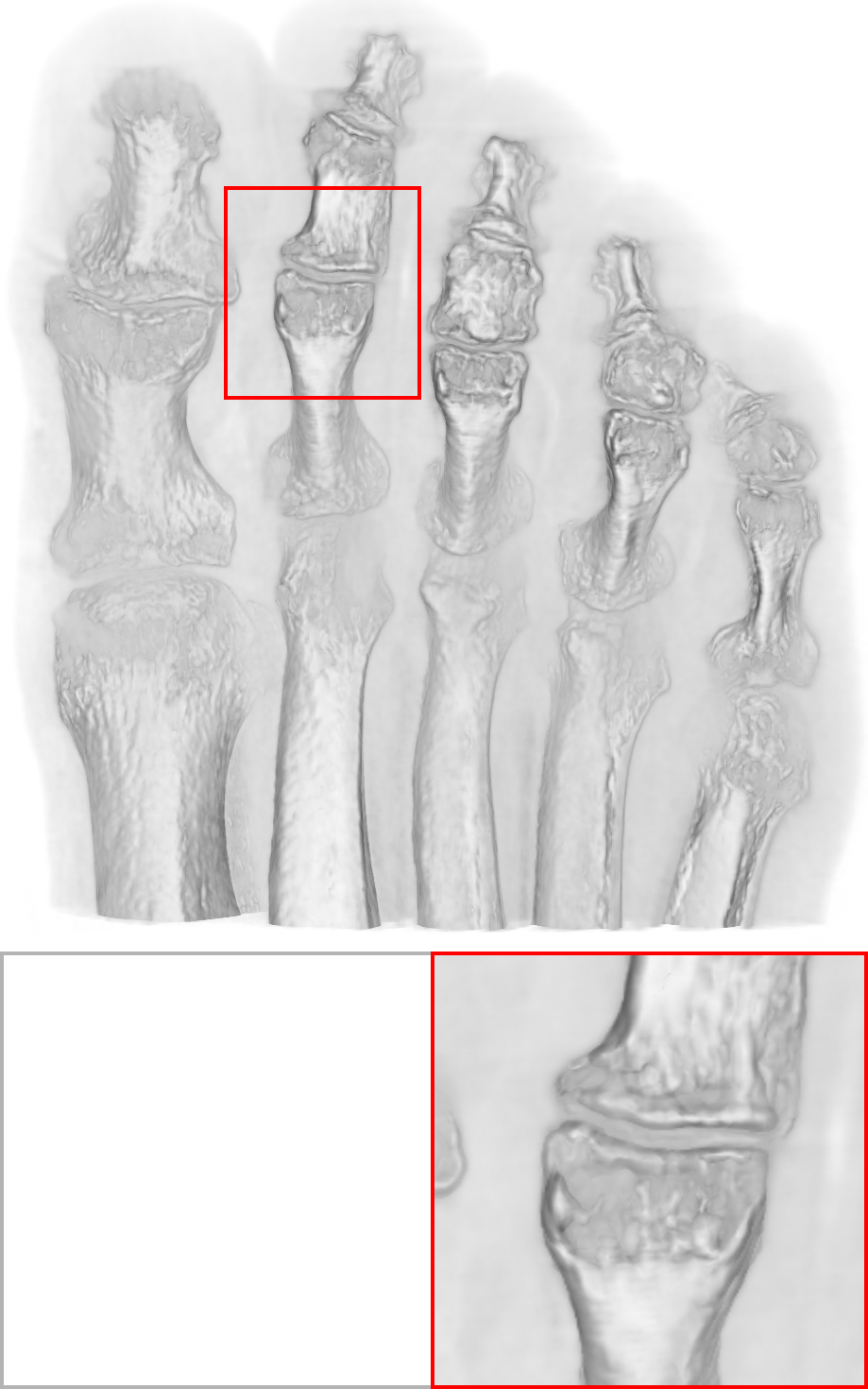} \\
    \includegraphics[width=0.1575\linewidth]{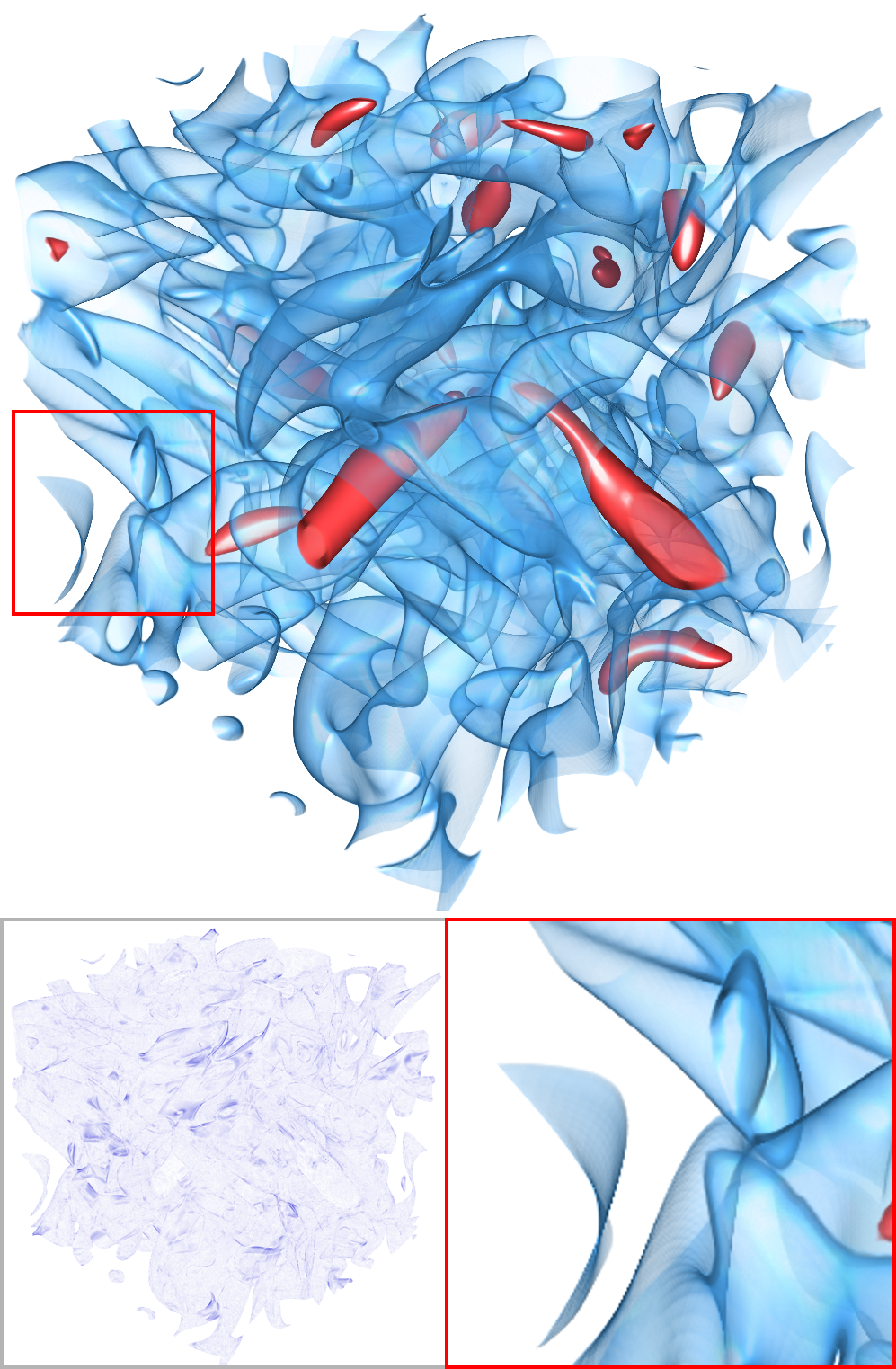} &
    \includegraphics[width=0.1575\linewidth]{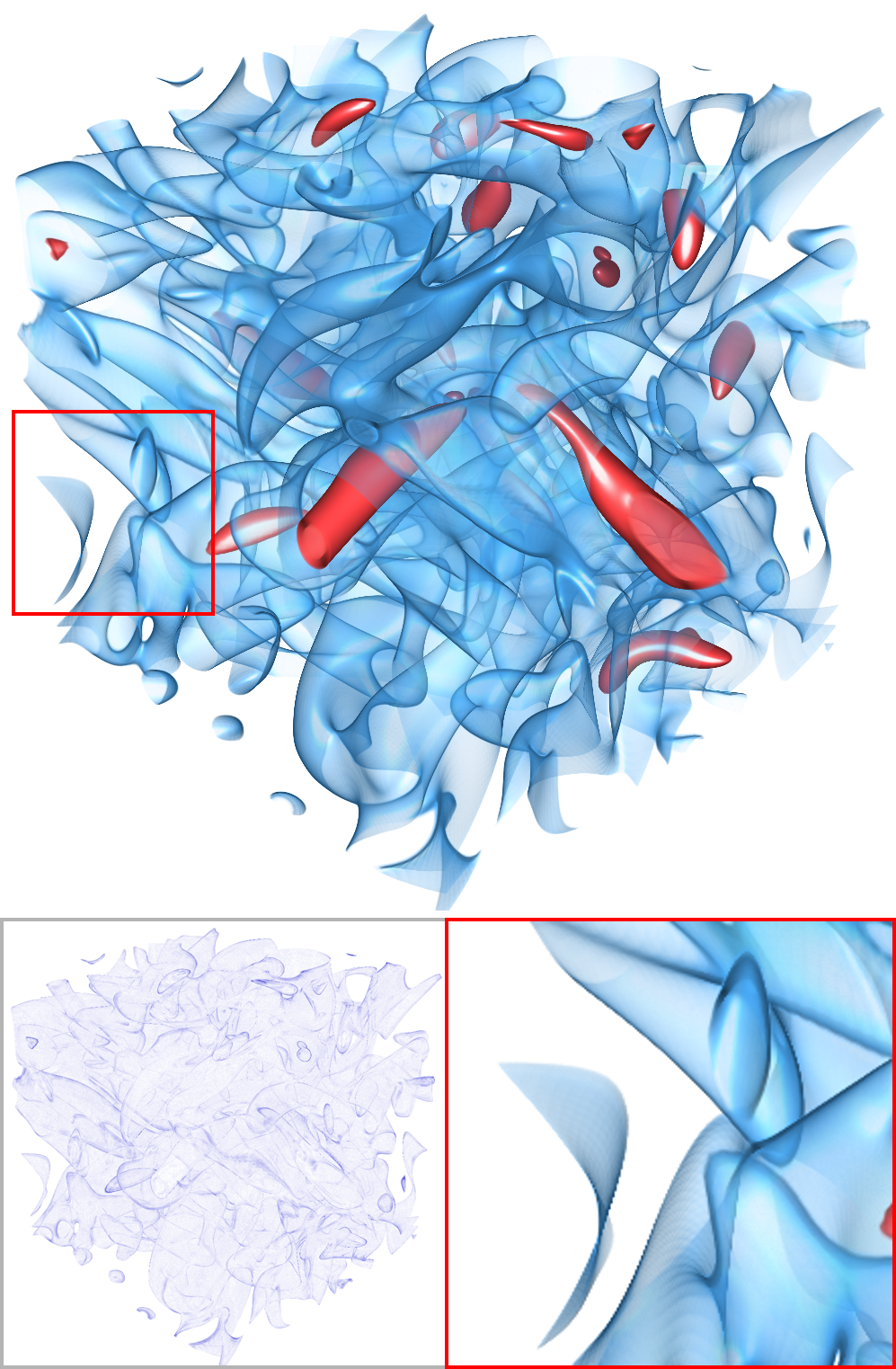} &
    \includegraphics[width=0.1575\linewidth]{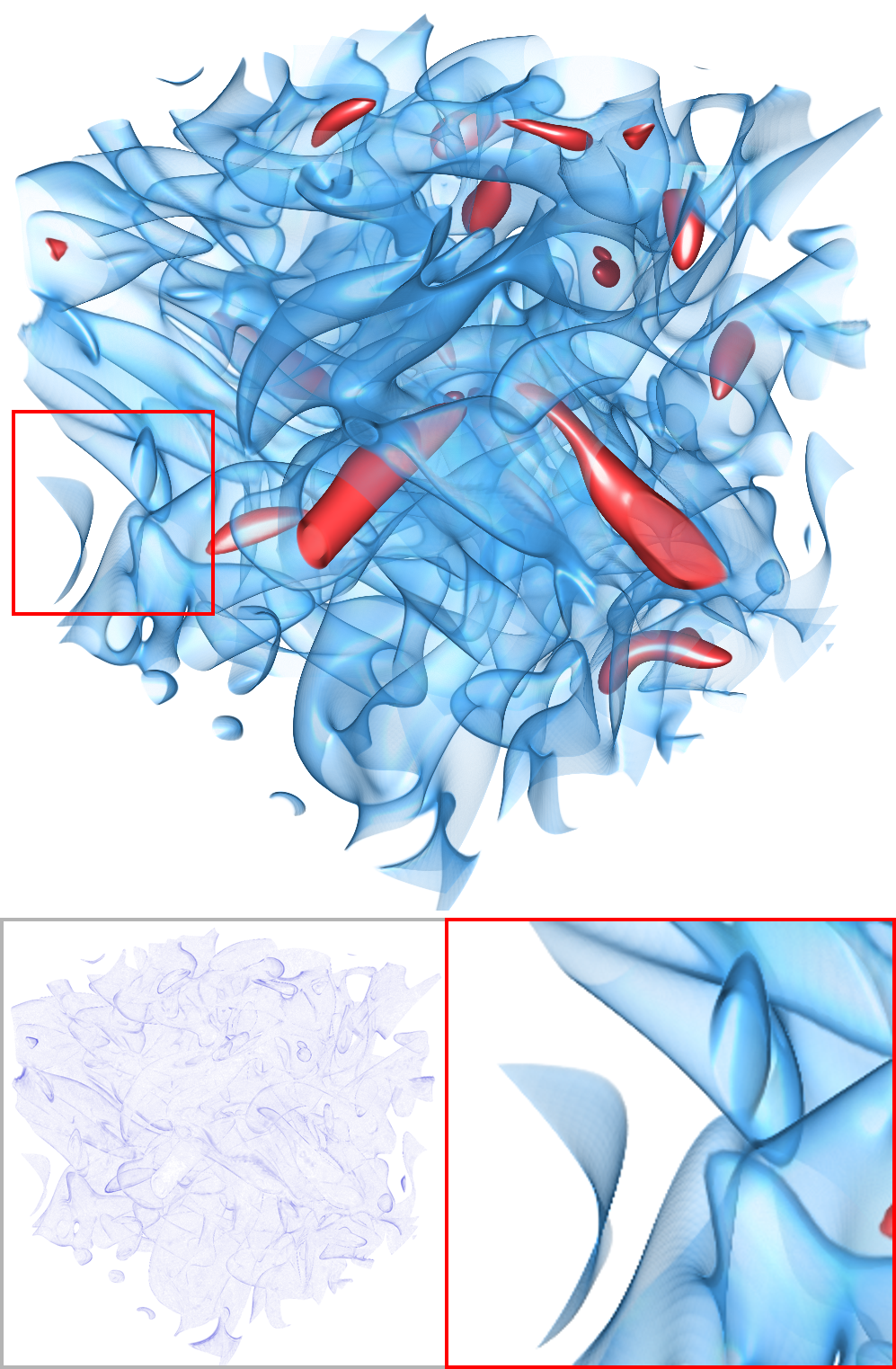} &
    \includegraphics[width=0.1575\linewidth]{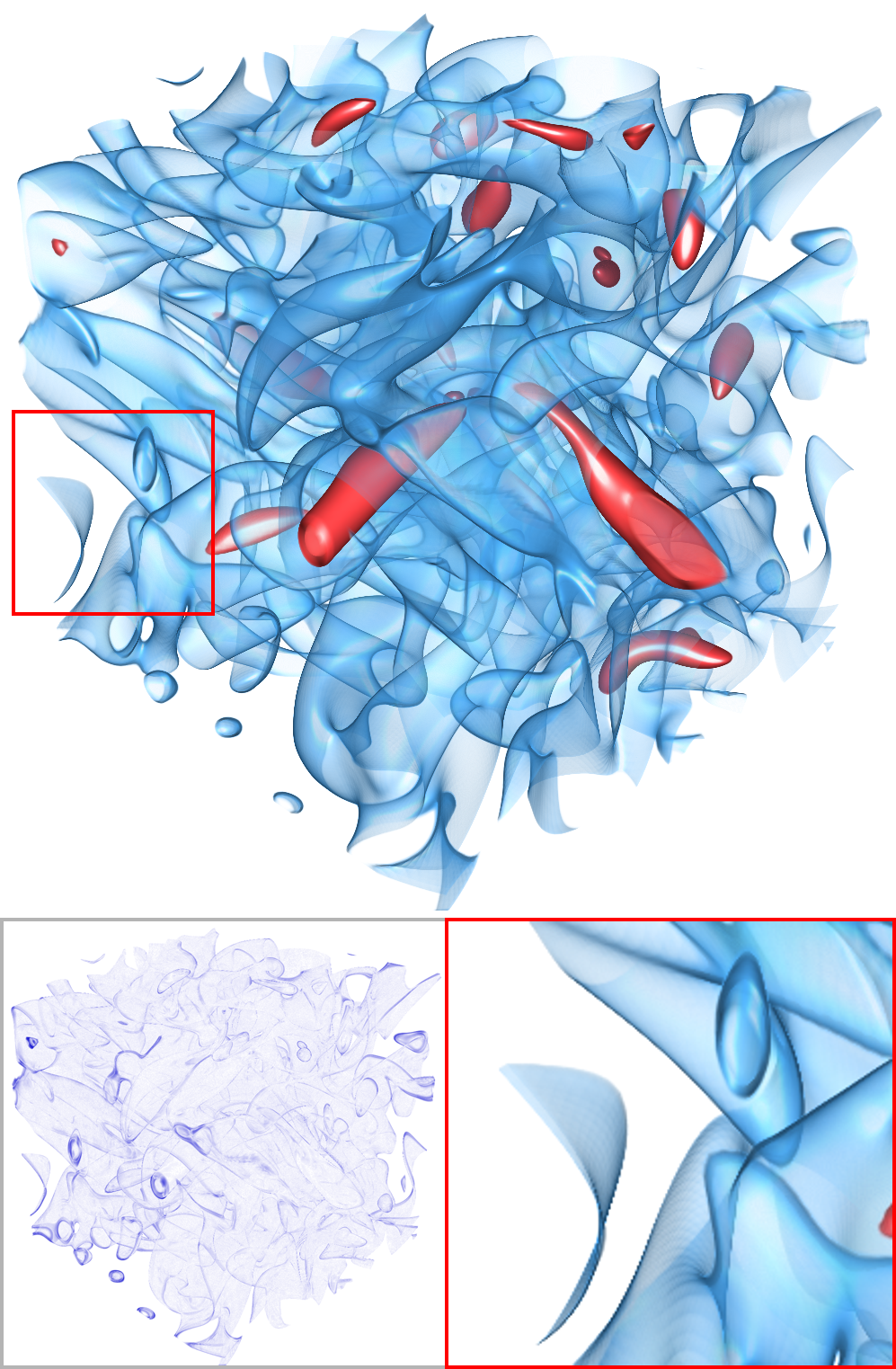} &
    \includegraphics[width=0.1575\linewidth]{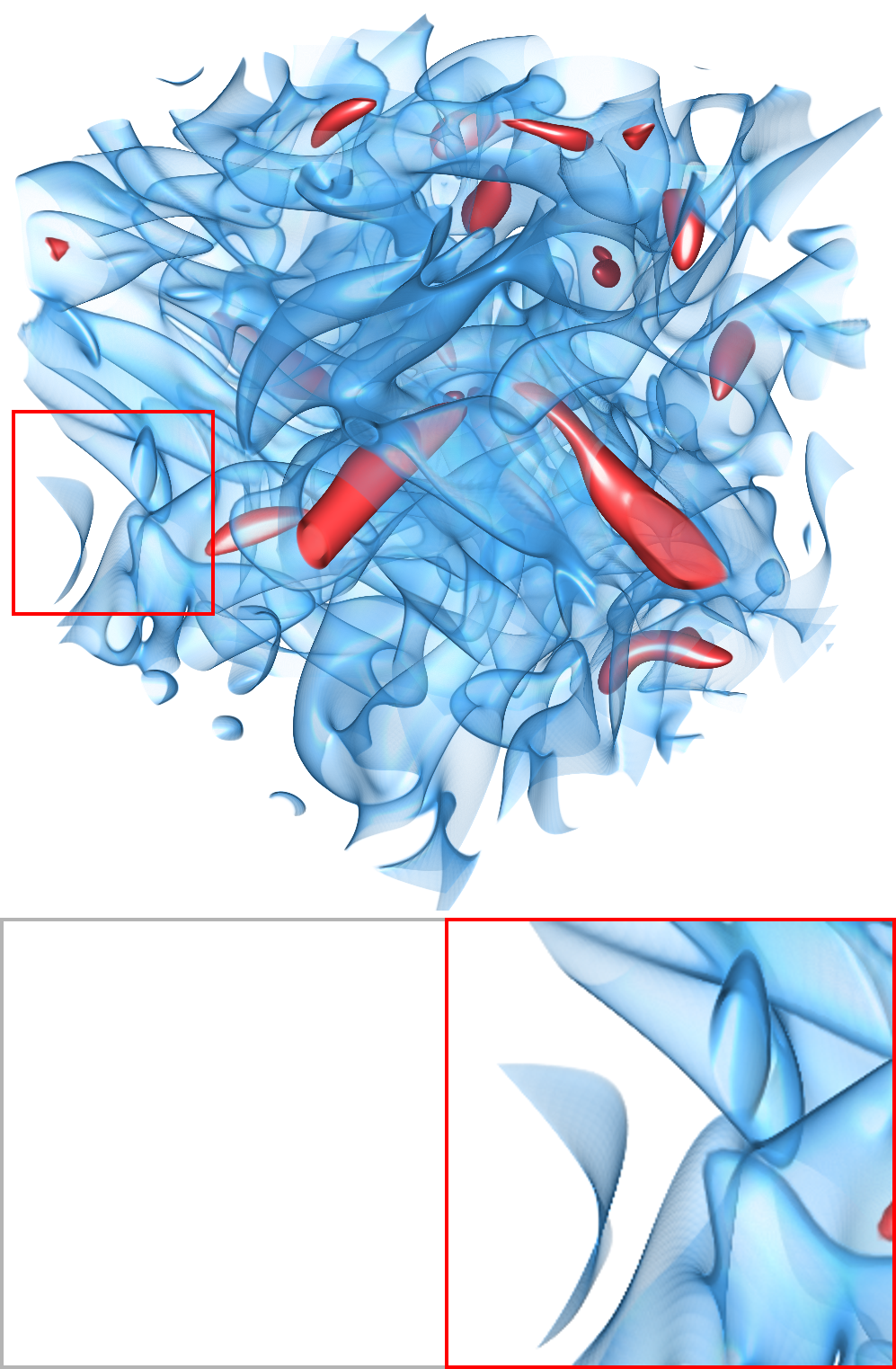} &
    \includegraphics[width=0.1575\linewidth]{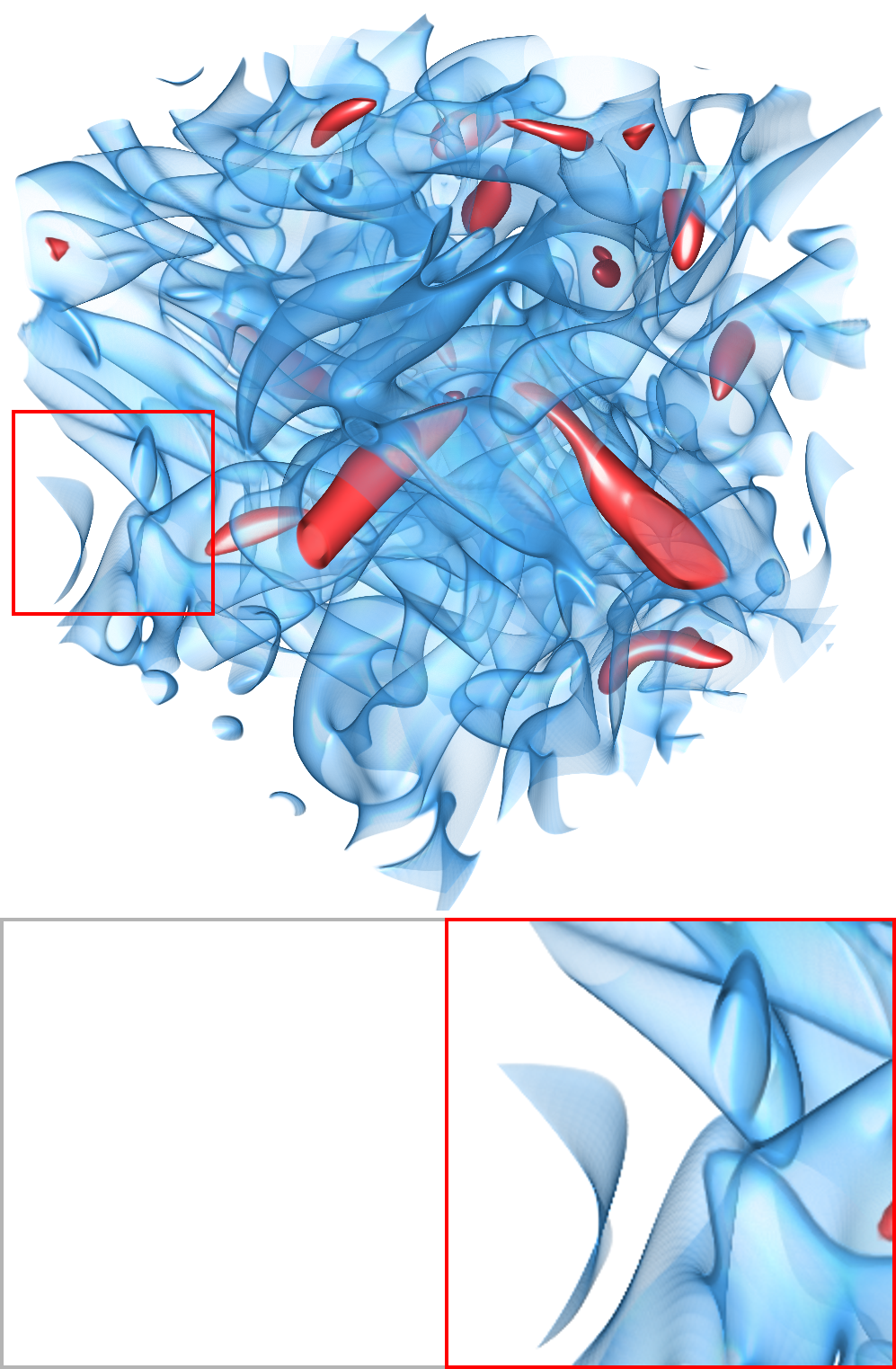} \\
    {\scriptsize ECNR} & {\scriptsize fV-SRN} & {\scriptsize Instant-NGP} & {\scriptsize AMGSRN++} & {\scriptsize Lossless-INR} & {\scriptsize ground truth} \\
\end{tabular}
\vspace{-.1in}
\caption{Comparing volume rendering results across different methods with the same model size. Top and bottom: \textsf{foot} and \textsf{vortex}. The bottom-left difference image shows pixel-wise error relative to the ground truth in the CIELUV color space, and the bottom-right inset is a zoom-in of the red-boxed region. Lossless-INR can losslessly reconstruct the ground-truth volume without error.}
\label{fig:baseline_comparison}
\vspace{-.1in}
\end{figure*}

\begin{figure}[!t]
\centering
\setlength{\tabcolsep}{2pt}
\renewcommand{\arraystretch}{1.0}
\begin{tabular}{@{}ccc@{}}
    \includegraphics[width=0.3\linewidth]{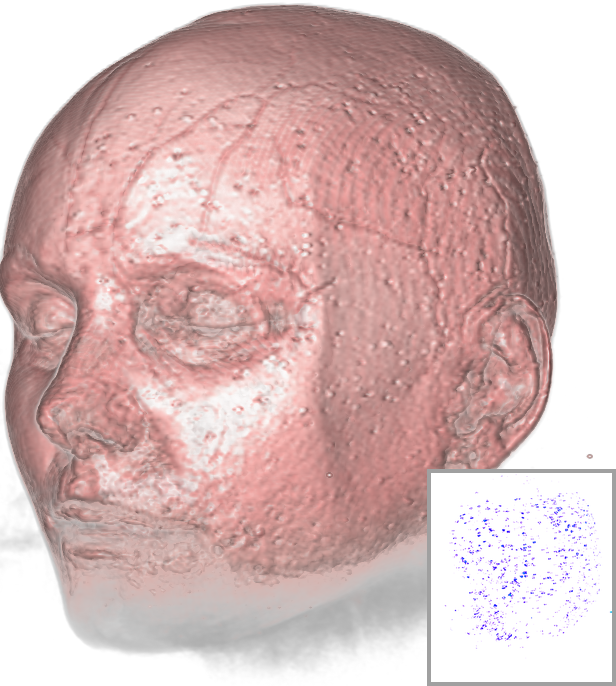} &
    \includegraphics[width=0.3\linewidth]{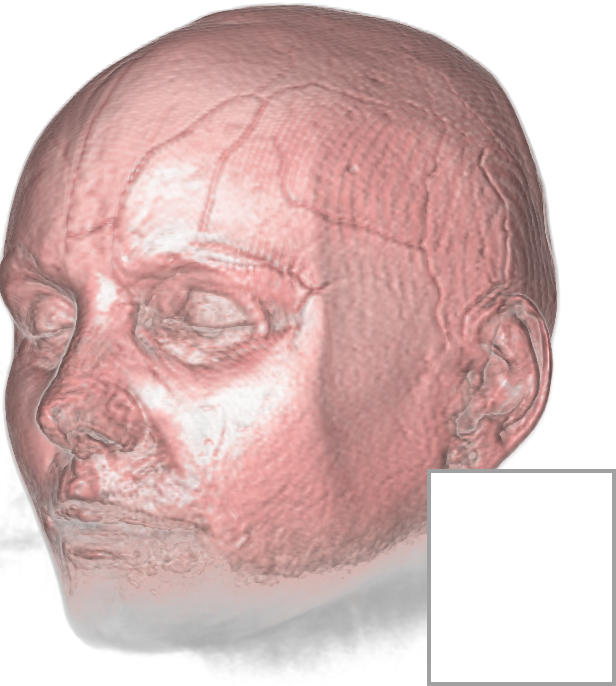} &
    \includegraphics[width=0.3\linewidth]{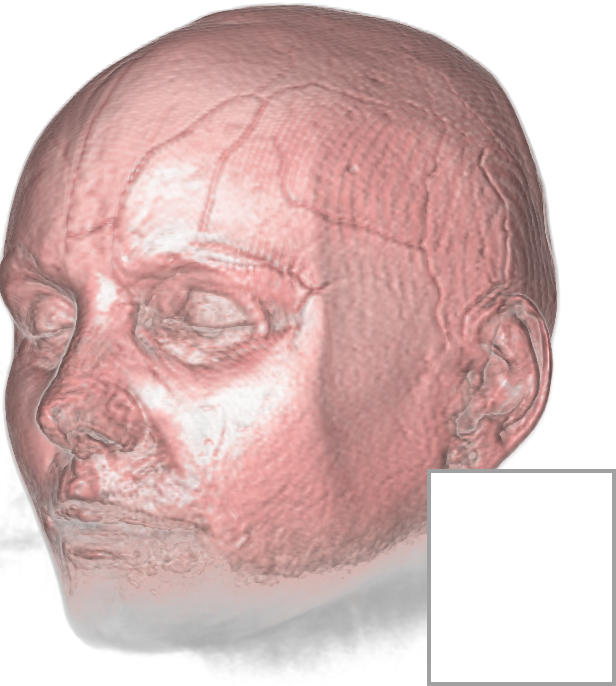} \\
    {\scriptsize binary grid} & {\scriptsize ternary grid (default)} & {\scriptsize floating-point grid} \\
\end{tabular}
\\[4pt]
\begin{tabular}{@{}cc@{}}
    \includegraphics[width=0.3\linewidth]{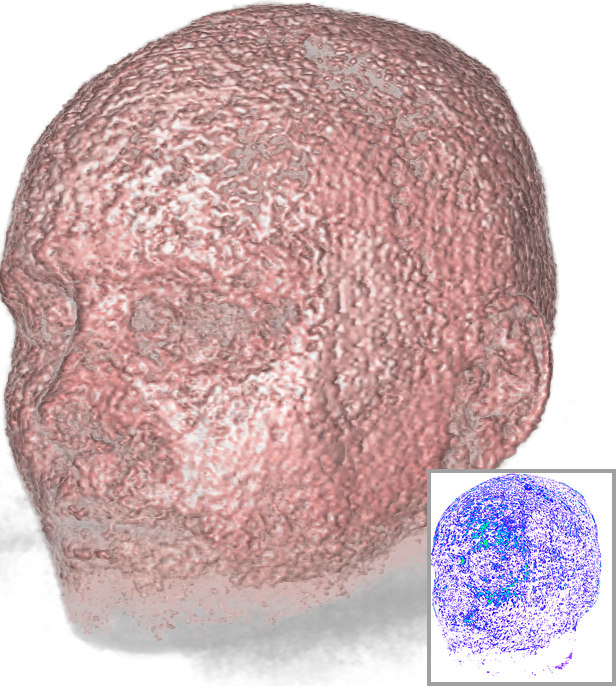} &
    \includegraphics[width=0.3\linewidth]{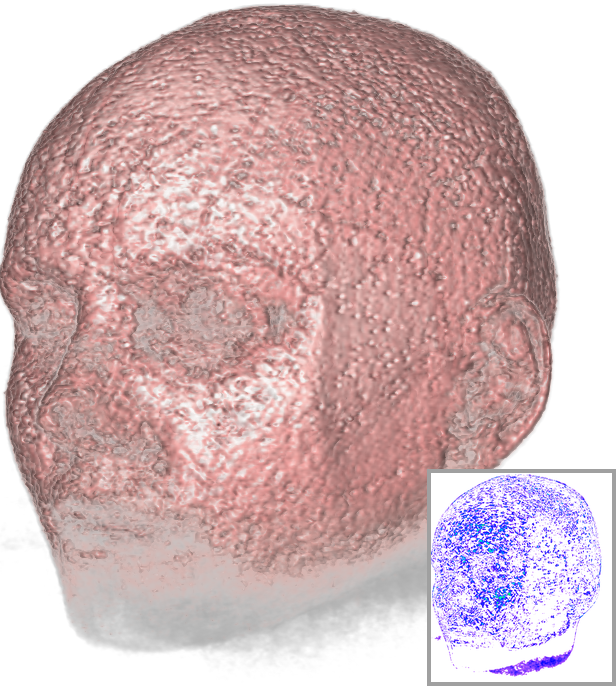} \\
    {\scriptsize w/o blocks + ternary grid} & {\scriptsize w/o blocks + floating-point grid} \\
\end{tabular}
\vspace{-.1in}
\caption{Volume rendering results across different model variants. Our default setting achieves lossless volumetric representation with a model size 20$\times$ smaller than its floating-point grid counterpart.}
\label{fig:ablation}
\vspace{-.1in}
\end{figure}

%% file: results.tex
\section{Results}
\label{sec:results}

\subsection{Datasets, Training, Baselines, and Metrics}

\textbf{Datasets and training.}
Table~\ref{tab:test_dataset} lists the volumetric datasets used in our experiments, all of which come from the Open SciVis Dataset repository~\cite{Klacansky-OpenScivis} and cover a range of resolutions and data types (8- and 16-bit unsigned integers and 32-bit floats). 
\hot{Due to page limit, we move the comparison of the \textsf{engine} and \textsf{tooth} datasets to Appendix~\ref{sec:appendix-tooth}.}
\pin{For each block, we use a 4-level multi-resolution hash grid with a hashmap size of $2^{17}$ and 2-dim features per entry together with four 3D feature grids, followed by a 3-layer, 64-hidden-unit MLP decoder.} We optimize with Adam for up to 2,000 iterations per block, using an initial learning rate of $10^{-2}$, cosine annealing to $10^{-5}$, and a batch size of 32,000. Training stops early once a block reaches zero bit errors. All experiments run on a workstation equipped with an NVIDIA 4090 GPU.

\textbf{Baselines and metrics.}
We compare against four representative INRs for volumetric data: ECNR~\cite{Tang-PacificVis24}, which fits each volume with a Laplacian pyramid of small parallelized block-wise MLPs; fV-SRN~\cite{Weiss-CGF22}, which combines a dense latent feature grid with Fourier-feature encoding for fast volume rendering; Instant-NGP~\cite{Muller-TOG22}, which couples a multi-resolution hash grid with a small MLP and shares the same backbone as our Lossless-INR, serving as the direct full-precision counterpart that isolates the effect of our bit-plane formulation and ternary parameterization; and AMGSRN++~\cite{Wurster-PacificVis25}, which adaptively places multiple feature grids according to local content complexity for large-scale volumetric data.
For a fair comparison, we set the baseline methods' model sizes to match those of Lossless-INR on each dataset.
We evaluate each method with three quality metrics. We assessed reconstruction quality using three complementary metrics: data-level {\em peak signal-to-noise ratio} (PSNR), image-level {\em learned perceptual image patch similarity} (LPIPS)~\cite{Zhang2018}, and bit-level {\em bit-error-rate} (BER), which is the ratio of incorrect bits of reconstructed volume to the total bits.

\vspace{-0.05in}
\subsection{Comparison with Baselines}

Table~\ref{tab:baseline_metrics} reports the quantitative results. The results show that existing lossy representation paradigms cannot achieve lossless reconstruction, even with large parameter budgets. However, because Lossless-INR sequentially encodes each block with the ternary feature grid network, its overall training time is substantially higher than that of the baseline methods. This effect is more pronounced on \textsf{foot} and \textsf{vortex}, where the higher data complexity requires more fine-level blocks to achieve lossless representation. Figure~\ref{fig:baseline_comparison} compares the volume-rendering results of different methods. Except for Lossless-INR, all baselines produce nonzero difference images because they do not achieve lossless reconstruction. This suggests that even when reconstruction fidelity is high, small errors can still accumulate during volume visualization, ultimately leading to noticeable deviations from the ground truth. In contrast, our Lossless-INR avoids this issue and is therefore well-suited to volume visualization and analysis tasks with low error tolerance.

\begin{table}[!t]
\centering
\caption{PSNR (dB), LPIPS, BER, and training time (TT, in minutes) of Lossless-INR and four INR baselines at matched model size.}
\label{tab:baseline_metrics}
\vspace{-0.05in}
\resizebox{1.0\columnwidth}{!}{%
\begin{tabular}{ccccccc}
\toprule
dataset & model size & method & PSNR $\uparrow$ & LPIPS $\downarrow$ & BER $\downarrow$ & TT \\
\midrule
  \multirow{5}{*}{\textsf{foot}} & \multirow{5}{*}{66.8\,MB} & ECNR & 36.27 & 0.0710 & 0.0942 & 2.07 \\
   &  & fV-SRN & 38.01 & 0.0368 & 0.0869 & 0.95 \\
   &  & Instant-NGP & 40.54 & 0.0229 & 0.0815 & 0.98 \\
   &  & AMGSRN++ & 39.41 & 0.0282 & 0.0814 & 4.47 \\
   &  & Lossless-INR & \bm{$+\infty$} & \textbf{0.0000} & \textbf{0.0000} & 22.20 \\
\midrule
  \multirow{5}{*}{\textsf{vortex}} & \multirow{5}{*}{91.9\,MB} & ECNR & 57.06 & 0.0058 & 0.2692 & 1.12 \\
   &  & fV-SRN & 57.29 & 0.0114 & 0.2678 & 1.55 \\
   &  & Instant-NGP & 60.46 & 0.0087 & 0.2650 & 1.63 \\
   &  & AMGSRN++ & 55.24 & 0.0178 & 0.2664 & 7.83 \\
   &  & Lossless-INR & \bm{$+\infty$} & \textbf{0.0000} & \textbf{0.0000} & 93.50 \\
\bottomrule
\end{tabular}
}
\vspace{-.1in}
\end{table}

\begin{table}[!t]
\centering
\caption{Ablation study results on the \textsf{MRI-woman} dataset.}
\label{tab:ablation}
\vspace{-0.075in}
\resizebox{0.85\columnwidth}{!}{%
\begin{tabular}{lccc}
\toprule
variant                                & lossless     & BER     & model size \\
\midrule
binary grid                            & $\times$     & $0.0029$  & 24.6\,MB \\
ternary grid (default)                          & $\checkmark$ & $0.0000$       & 39.2\,MB \\
floating-point grid                             & $\checkmark$ & $0.0000$       & 794.4\,MB \\
w/o blocks + ternary grid           & $\times$     & $0.1989$  & 1.2\,MB \\
w/o blocks + floating-point grid             & $\times$     & $0.1861$  & 24.9\,MB \\
\bottomrule
\end{tabular}
}
\vspace{-.1in}
\end{table}

\vspace{-0.05in}
\subsection{Ablation Studies}

In this section, we investigate how the parameter precision of feature grid and octree block partitioning can influence the lossless representation.
On the \textsf{MRI-woman} dataset, we compare three feature-grid precisions: binary~\cite{Shin-NeurIPS23} ($\{-1,+1\}$), ternary ($\{-1,0,+1\}$), and full floating-point precision~\cite{Muller-TOG22}, as well as two variants that fit the entire volume with a single INR without block partitioning. 
Table~\ref{tab:ablation} reports, for each variant, whether it achieves lossless reconstruction, the bit-error rate, and the resulting model size.
In addition, Figure~\ref{fig:ablation} provides a qualitative comparison of the reconstructed volumes across all variants.
Specifically, the binary grid yields the most compact block-wise model but cannot drive its BER to zero, thereby failing to achieve a lossless representation.
The full-precision grid also achieves lossless reconstruction, but its storage cost is approximately 20 times that of the ternary grid. 
Without block partitioning, neither the floating-point grid nor the ternary grid achieves lossless reconstruction, and both variants exhibit high BER. 
This highlights the necessity of octree block partitioning for lossless representation.

%% file: conclusions.tex
\section{Conclusions and Future Work}
\label{sec:conclusions}

We have explored the feasibility of Lossless-INR, a lossless volumetric implicit neural representation. Through bit-plane decomposition and octree block partitioning, Lossless-INR can achieve lossless representation with limited network capacity. To reduce the final model size, we design a ternary feature grid network to encode each partitioned block, whose grid weights are constrained to a ternary set, enabling lossless representation while reducing storage cost by 20$\times$ compared with the full-precision counterpart.

However, the model size produced by Lossless-INR after encoding a volume remains relatively large, \hot{and can even be larger than the original volume. Although this paper focuses on volume representation rather than compression, our future work will aim to further reduce parameter storage costs while preserving nearly the same representational capacity.} One promising direction is to incorporate deep compression techniques~\cite{Han-ICLR16}, such as network pruning~\cite{Cheng-TPAMI24} and quantization-aware training~\cite{Jacob-CVPR18}. Another limitation is the long optimization time of the current framework. \hot{Since blocks are optimized sequentially, the training cost grows with the number of blocks, which hinders scaling Lossless-INR to large-scale volumes where optimization could take several days.}
A straightforward solution is to train all blocks in parallel with multiple GPUs, although this would increase the required computational resources. \hot{The per-block optimization can also be accelerated, e.g., by a meta-learned initialization~\cite{Yang-PacificVis25} shared across blocks. Hence, each converges in fewer iterations, or by engineering optimizations such as mixed-precision training and fully-fused MLP kernels.} Overall, lossless representation is a promising direction for future research, and more compact, efficient models built on a lossless foundation will offer greater practical value.

%% file: appendix.tex
%

\newpage
\clearpage

\setcounter{section}{0}
\setcounter{figure}{0}
\setcounter{table}{0}

\section{\hot{Comparing on Additional Datasets}}
\label{sec:appendix-tooth}

\begin{figure}[htb]
\centering
\setlength{\tabcolsep}{1pt}
\renewcommand{\arraystretch}{1.0}
\begin{tabular}{@{}cccccc@{}}
    \includegraphics[width=0.159\linewidth]{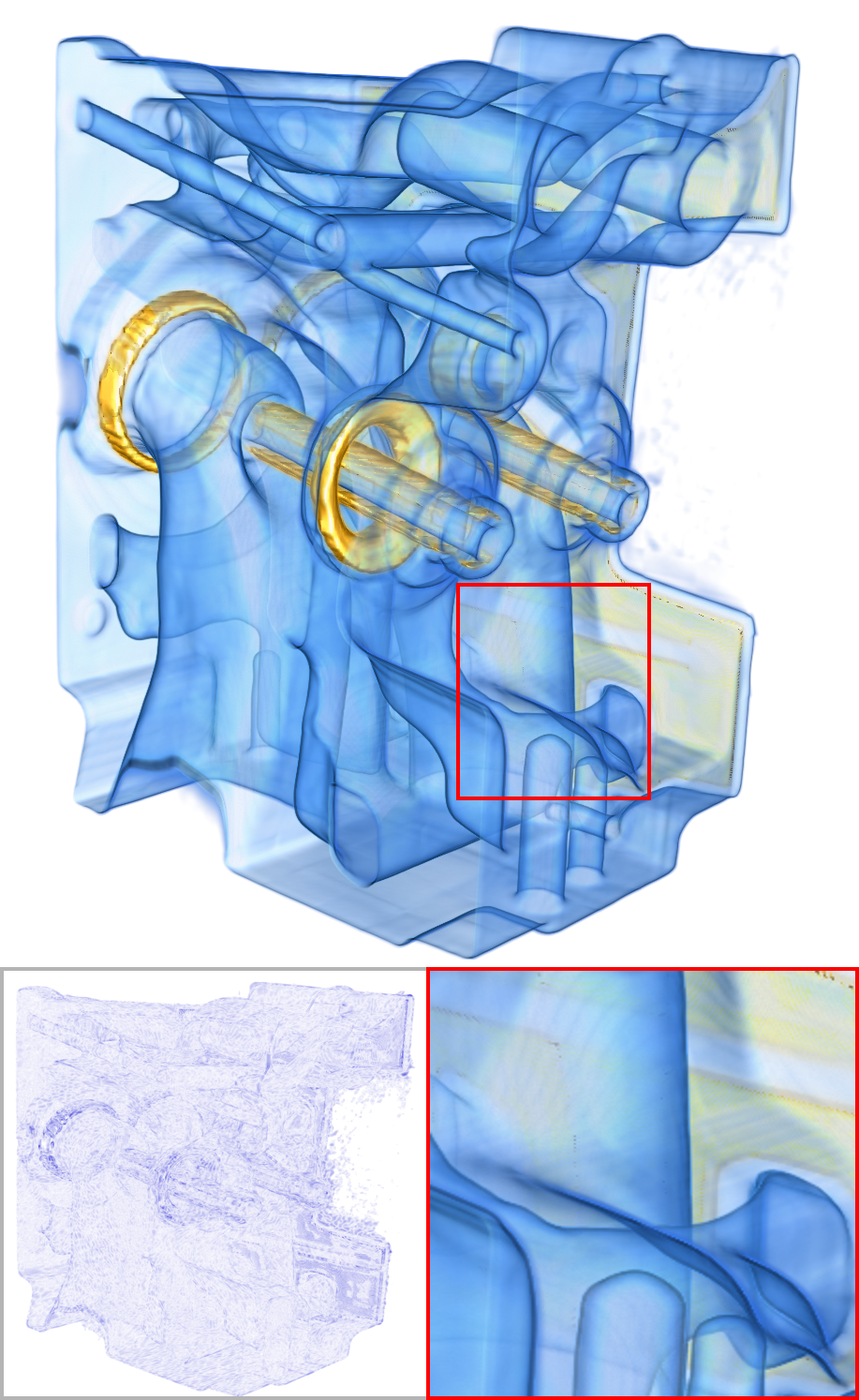} &
    \includegraphics[width=0.159\linewidth]{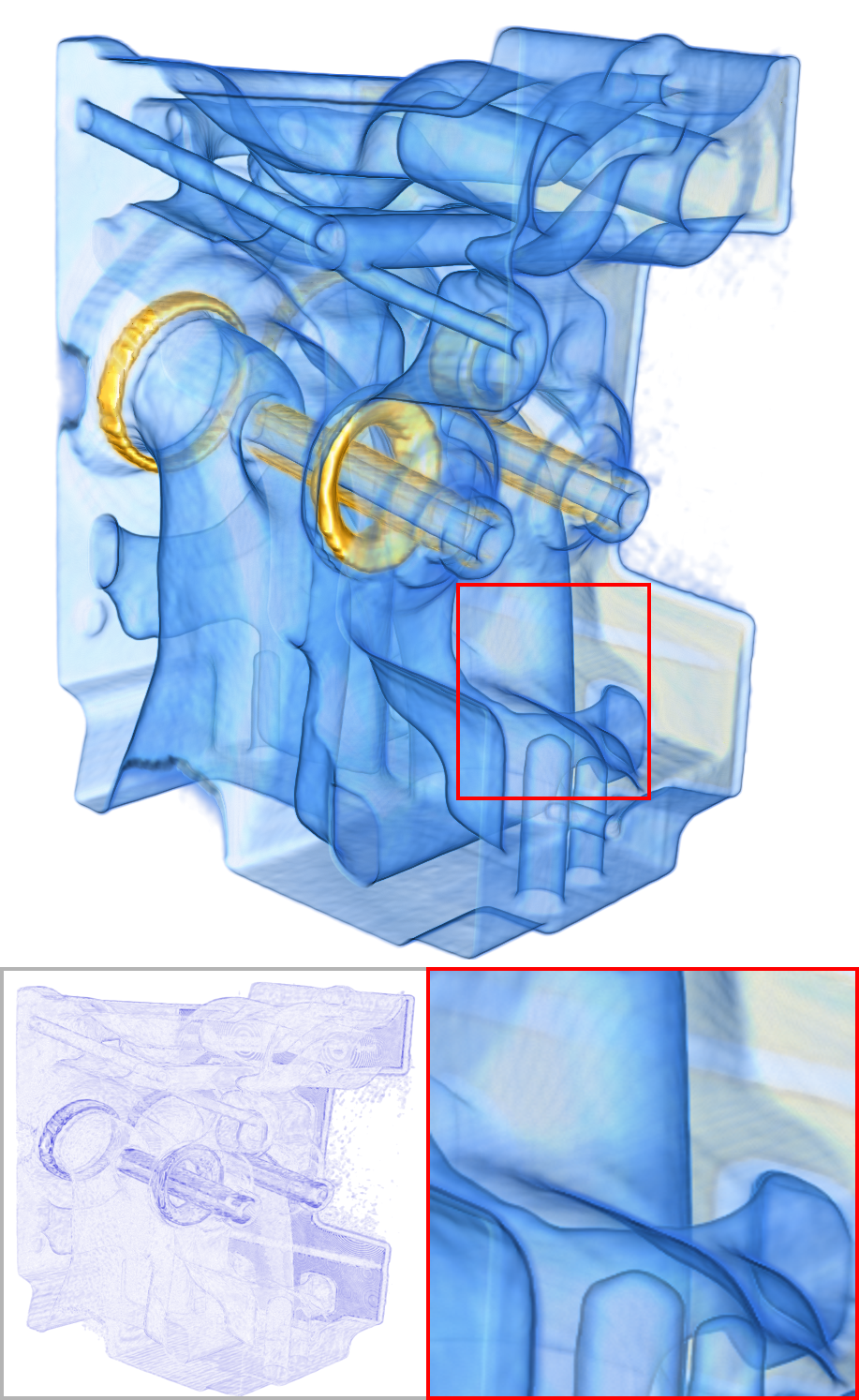} &
    \includegraphics[width=0.159\linewidth]{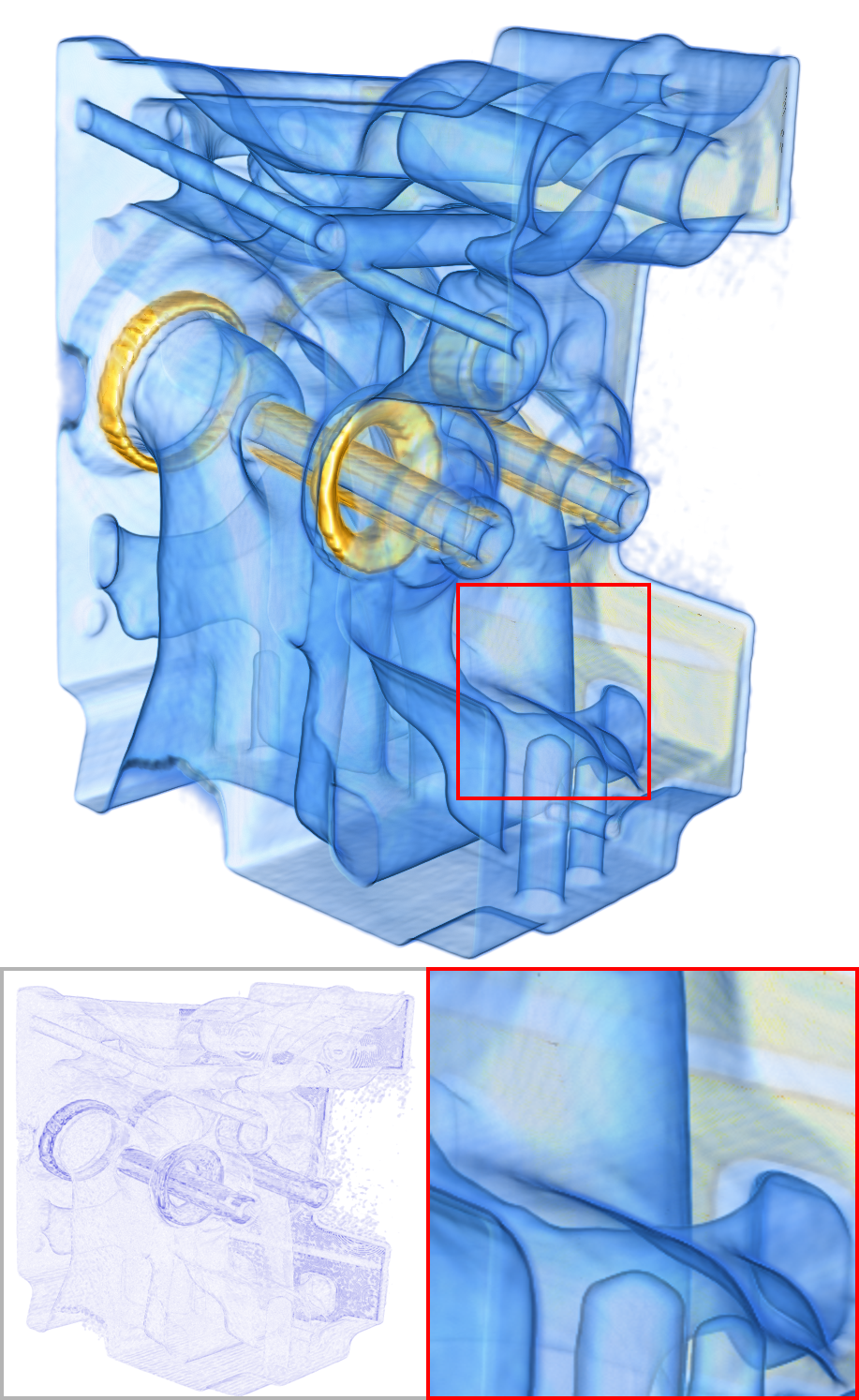} &
    \includegraphics[width=0.159\linewidth]{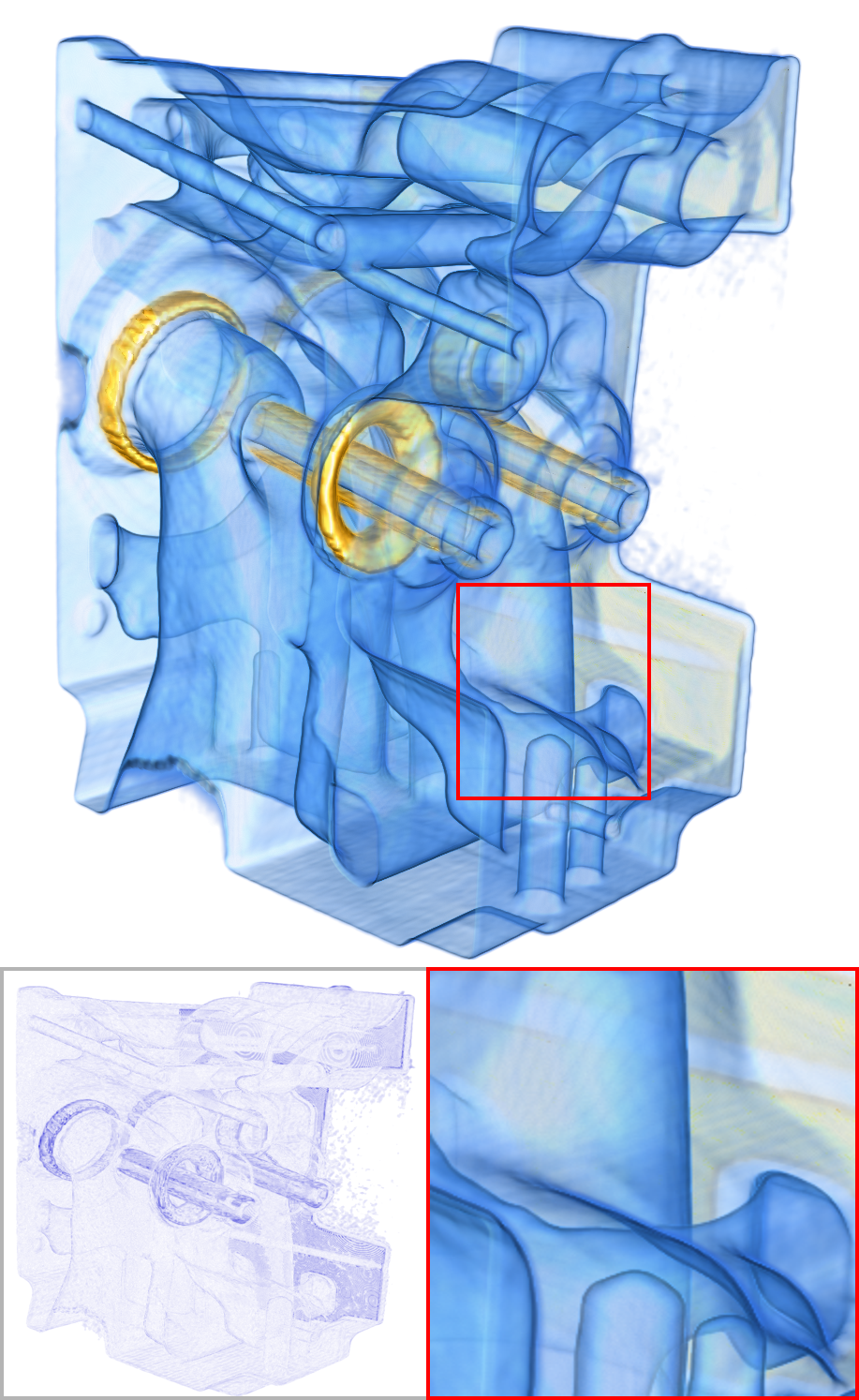} &
    \includegraphics[width=0.159\linewidth]{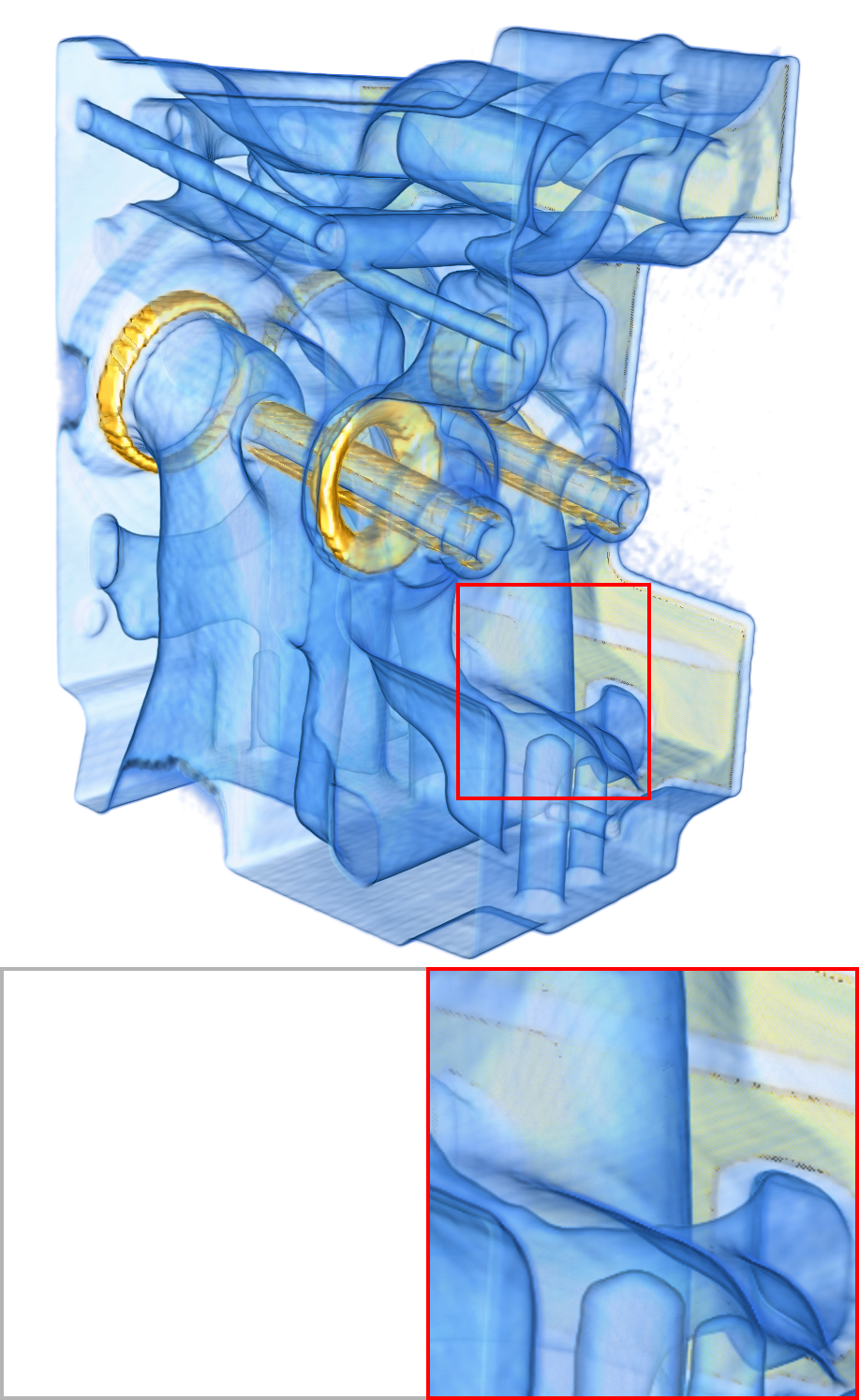} &
    \includegraphics[width=0.159\linewidth]{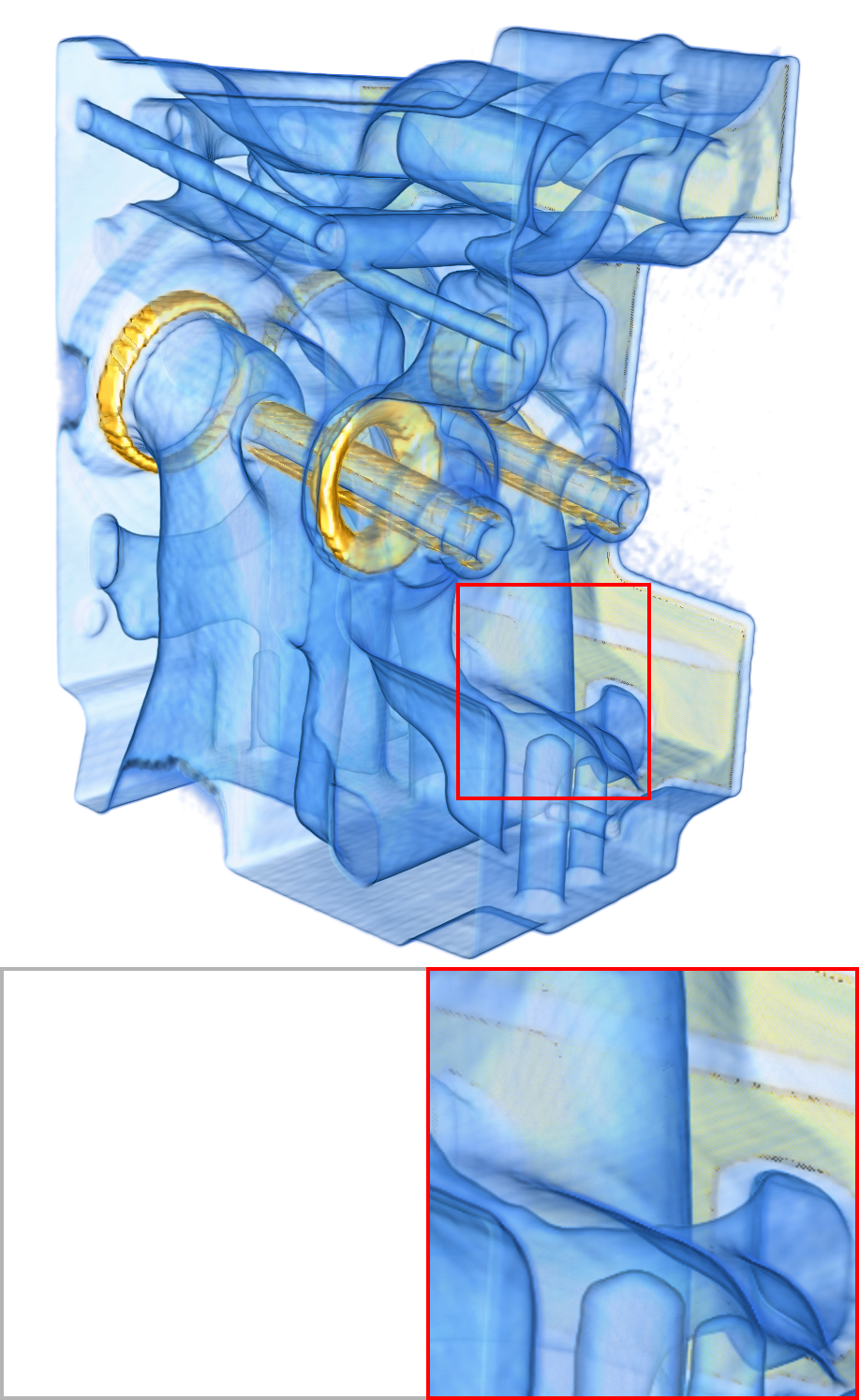} \\
    \includegraphics[width=0.159\linewidth]{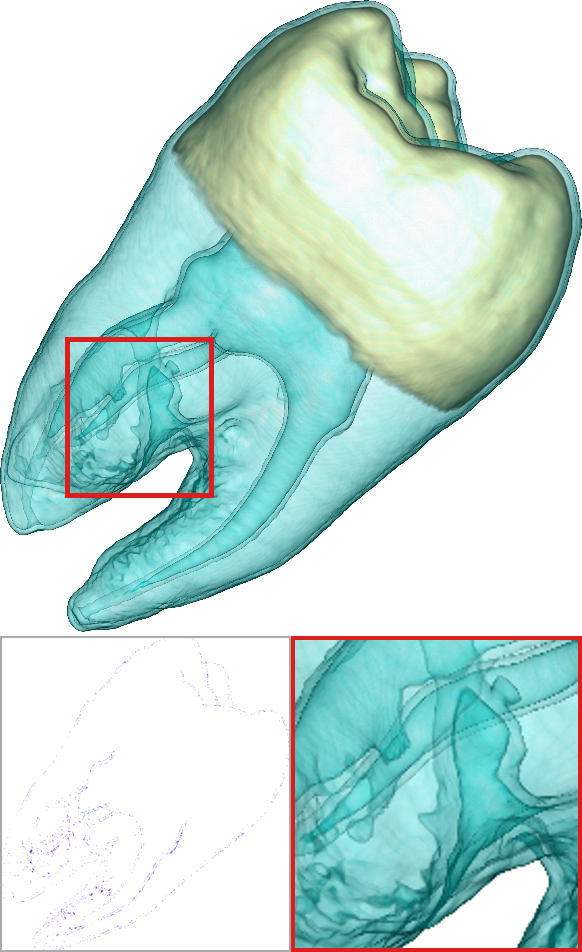} &
    \includegraphics[width=0.159\linewidth]{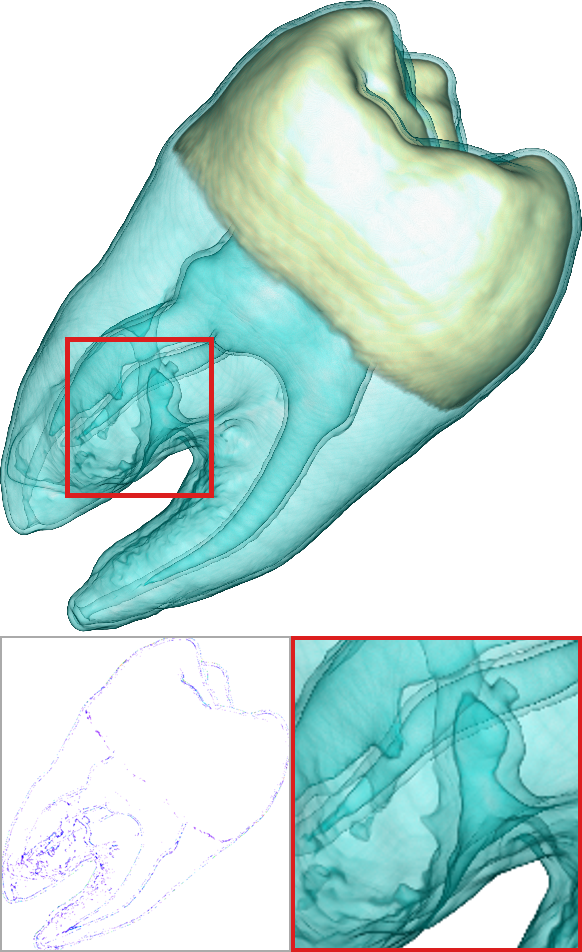} &
    \includegraphics[width=0.159\linewidth]{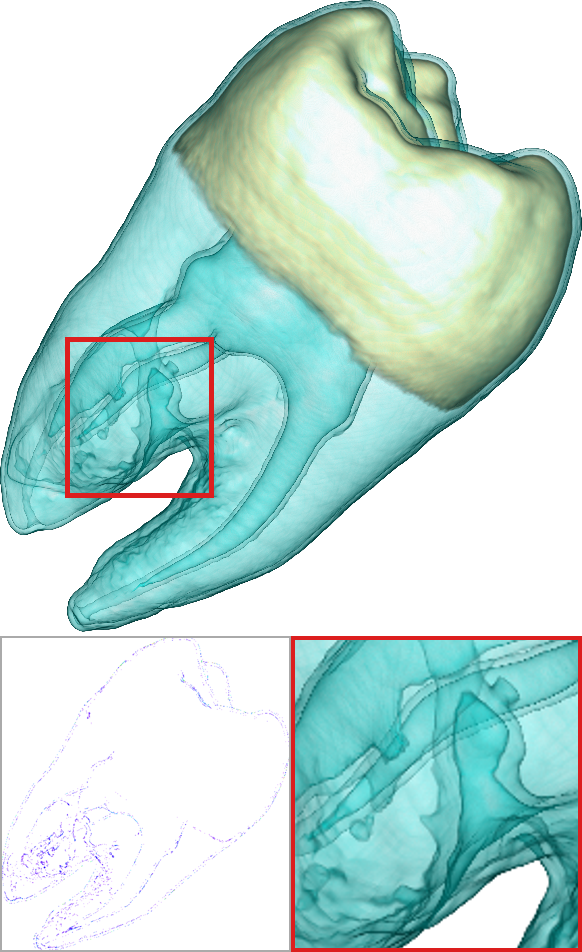} &
    \includegraphics[width=0.159\linewidth]{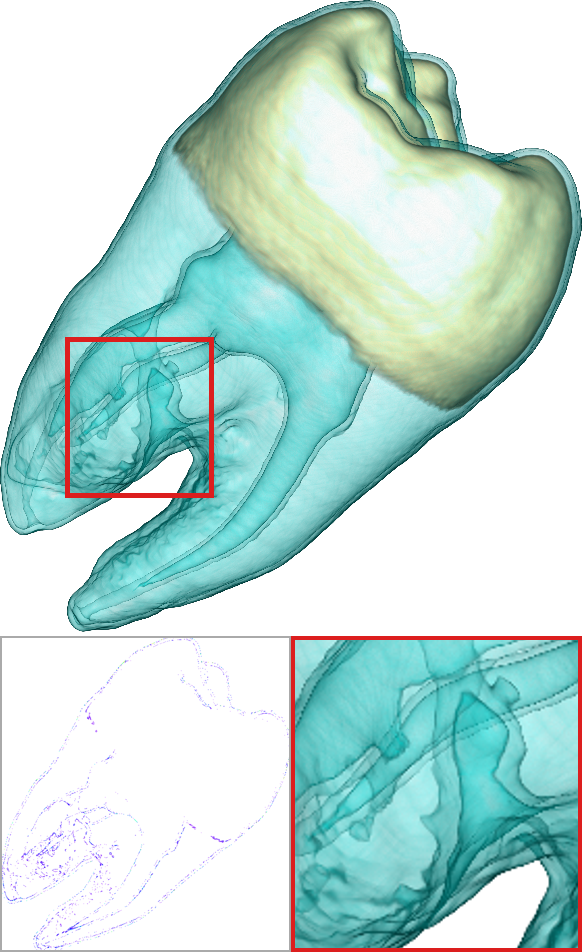} &
    \includegraphics[width=0.159\linewidth]{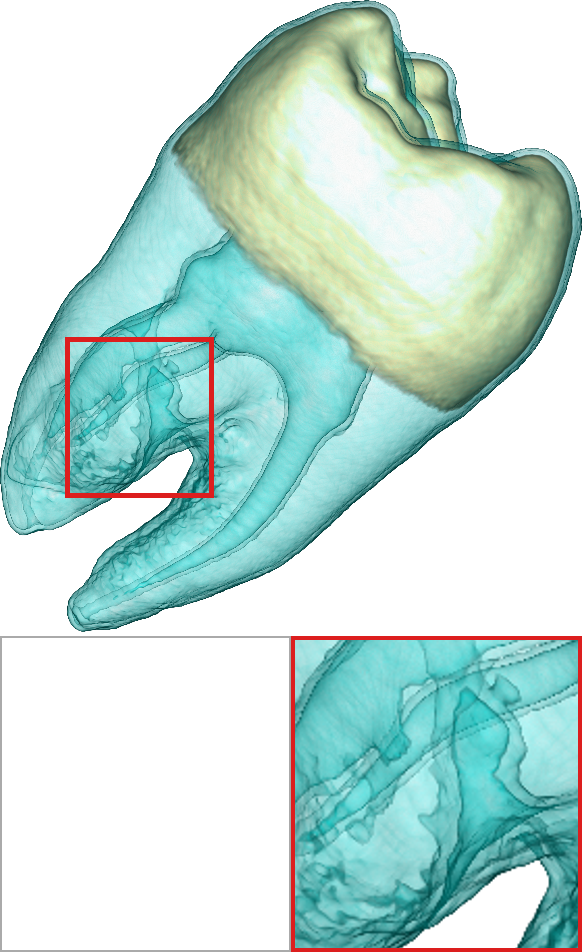} &
    \includegraphics[width=0.159\linewidth]{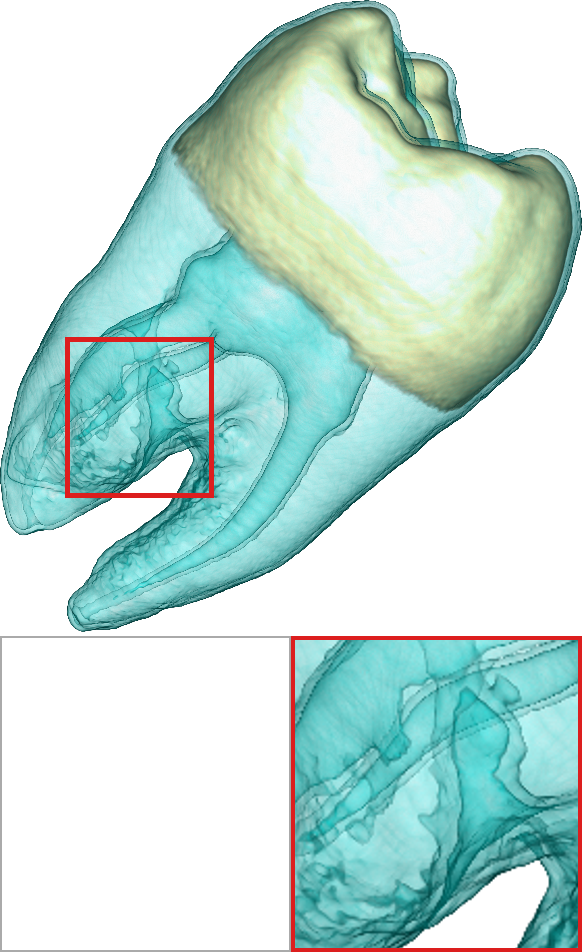} \\
    {\small ECNR} & {\small fV-SRN} & {\small Instant-NGP} & {\small AMGSRN++} & {\small Lossless-INR} & {\small ground truth} \\
\end{tabular}
\vspace{-.1in}
\caption{\hot{
Comparing volume rendering results across different methods with the same model size. Top and bottom: \textsf{engine} and \textsf{tooth}. The bottom-left difference image shows pixel-wise error relative to the ground truth in the CIELUV color space, and the bottom-right inset is a zoom-in of the red-boxed region. 
}
}
\label{fig:appendix_comparison}
\end{figure}

\hot{Due to page limit, we do not include all the comparisons in the main paper. 
In this section, we report the results for the other two datasets, \textsf{engine} and \textsf{tooth}.
As in the comparison in the main paper, we keep the model sizes for all methods the same.
Table~\ref{tab:appendix_metrics} lists the quantitative results, and Figure~\ref{fig:appendix_comparison} shows the corresponding volume rendering results.
We can observe that for lossy representation methods, even when reconstruction quality is high, the reconstruction still contains errors in a non-blank difference image. 
In contrast, Lossless-INR is the only method that achieves a zero error rate and produces pixel-wise identical results to the ground truth.}

\begin{table}[htb]
\centering
\caption{\hot{PSNR (dB), LPIPS, BER, and training time (TT, in minutes) of Lossless-INR and four INR baselines on the \textsf{engine} and \textsf{tooth} datasets at matched model size.}}
\label{tab:appendix_metrics}
\vspace{-0.05in}\hot{
\resizebox{0.9\columnwidth}{!}{%
\begin{tabular}{ccccccc}
\toprule
dataset & model size & method & PSNR $\uparrow$ & LPIPS $\downarrow$ & BER $\downarrow$ & TT \\
\midrule
  \multirow{5}{*}{\textsf{engine}} & \multirow{5}{*}{61.0\,MB} & ECNR & 45.67 & 0.0446 & 0.0728 & 1.82 \\
   &  & fV-SRN & 47.10 & 0.0446 & 0.0682 & 0.93 \\
   &  & Instant-NGP & 49.99 & 0.0311 & 0.0606 & 0.98 \\
   &  & AMGSRN++ & 48.47 & 0.0332 & 0.0647 & 4.63 \\
   &  & Lossless-INR & \bm{$+\infty$} & \textbf{0.0000} & \textbf{0.0000} & 2.20 \\
\midrule
  \multirow{5}{*}{\textsf{tooth}} & \multirow{5}{*}{12.5\,MB} & ECNR & 49.68 & 0.0443 & 0.0888 & 0.91 \\
   &  & fV-SRN & 46.76 & 0.0720 & 0.1578 & 0.43 \\
   &  & Instant-NGP & 47.36 & 0.0571 & 0.1601 & 0.70 \\
   &  & AMGSRN++ & 46.02 & 0.0527 & 0.1631 & 1.16 \\
   &  & Lossless-INR & \bm{$+\infty$} & \textbf{0.0000} & \textbf{0.0000} & 0.94 \\
\bottomrule
\end{tabular}
}}
\end{table}

\section{\hot{Rate-Distortion Curves}}
\label{sec:appendix-sweep}

\begin{figure}[!t]
\centering
\setlength{\tabcolsep}{2pt}
\begin{tabular}{@{}cc@{}}
    \includegraphics[height=1.45in]{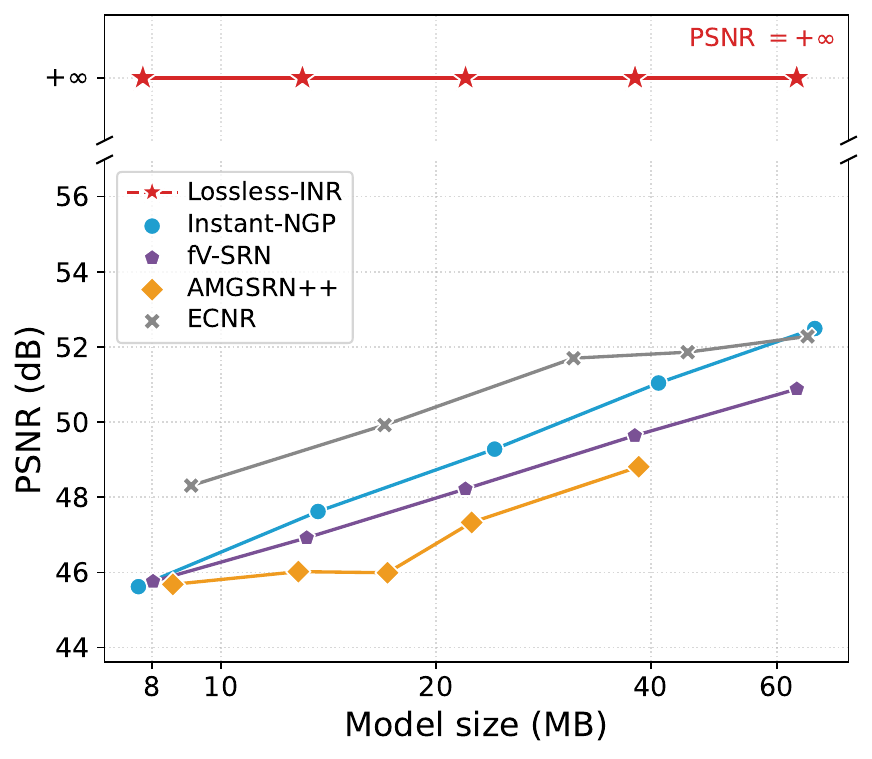} &
    \includegraphics[height=1.45in]{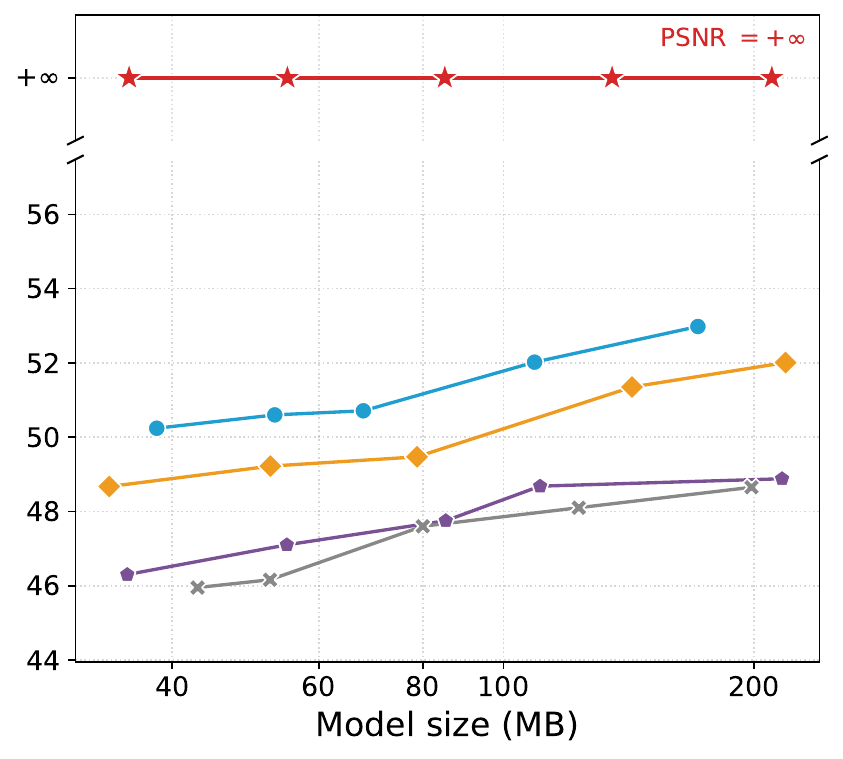} \\
    {\scriptsize (a) \textsf{tooth}} & {\scriptsize (b) \textsf{engine}} \\
\end{tabular}
\vspace{-.1in}
\caption{\hot{Rate-distortion curves for different methods on two datasets.}}
\label{fig:sweep}
\end{figure}

\hot{We report the rate-distortion behavior of each method in Figure~\ref{fig:sweep}, where data-level PSNR is plotted with varying model sizes on the \textsf{tooth} and \textsf{engine} datasets. This analysis goes beyond a single comparison at a matched size and provides practical guidance on when a lossless representation becomes necessary.}
\hot{Conventional INR baselines generally improve reconstruction quality by increasing the model size, since additional parameters increase representational capacity. 
However, these methods still optimize a voxel-value regression objective. As a result, their reconstruction quality may eventually saturate and, in some cases, even degrade once the model size exceeds a certain threshold; adding parameters yields only trivial improvements in quality. Therefore, further improving fidelity requires changing the formulation rather than merely increasing capacity. Lossless-INR addresses this limitation by predicting the per-bit values associated with each voxel, which enables bit-exact reconstruction. From this perspective, the model size at which Lossless-INR achieves lossless reconstruction can be interpreted as an upper bound on the parameter budget that is worth spending on a lossy representation. For example, in Figure~\ref{fig:sweep} (a), Lossless-INR already achieves bit-exact reconstruction at roughly $8$ MB. Once the baseline methods exceed this size without achieving lossless fidelity, their voxel-value objective becomes the limiting factor, indicating that one should switch to Lossless-INR rather than further scaling the lossy models. We note that the current Lossless-INR model is still larger than the raw volume. Nevertheless, we believe future designs can substantially reduce this overhead while preserving the random-access advantage of INR representations, ultimately enabling efficient volume visualization.}

